%% file: main.tex
\documentclass[journal]{IEEEtran}

\input{usepackage}

\begin{document}
%
\title{GPL-SLAM: A Laser SLAM Framework with Gaussian Process Based Extended Landmarks}
%
%


\author{
    \IEEEauthorblockN{Ali Emre Balcı \IEEEauthorrefmark{1} \IEEEauthorrefmark{2} \thanks{\IEEEauthorrefmark{2} Authors contributed equally}, Erhan Ege Keyvan \IEEEauthorrefmark{4} \IEEEauthorrefmark{2}, Emre Özkan \IEEEauthorrefmark{4}} \\
    \IEEEauthorblockA{\IEEEauthorrefmark{1} TU Delft
        \\ aebalci@tudelft.nl} \\
    \IEEEauthorblockA{\IEEEauthorrefmark{4} Middle East Technical University
        \\ \{ege.keyvan, emreo\}@metu.edu.tr}
}

\maketitle

\input{sections/abstract}

%
\IEEEpeerreviewmaketitle

\input{sections/intro}

\input{sections/method}

\input{sections/results}

\input{sections/discussion}

\input{sections/conclusion}

\input{sections/appendix}

\input{sections/references}


\end{document}

%% file: usepackage.tex
\usepackage{amsmath, amssymb, amsthm, bm}
\usepackage{graphicx}
\usepackage{cite}
\usepackage{url}
\usepackage{tikz}
\usepackage{standalone}
\usepackage{subcaption}
\usepackage{xcolor}
\usepackage[utf8]{inputenc}
\usepackage{booktabs}
\usepackage{mathtools}
\usepackage{array}
\usepackage{xpatch}
\usepackage{floatrow}
\usepackage[linktocpage,hidelinks]{hyperref}
\usepackage{algorithm}
\usepackage{algpseudocodex}
\input{math_defn}

%% file: math_defn.tex
\newcommand{\R}{\mathbb{R}}

\newcommand{\bx}{\bm{x}}
\newcommand{\bSig}{\bm{\Sigma}}
\newcommand{\bv}{\bm{v}}
\newcommand{\bz}{\bm{z}}

\newcommand{\bR}{\bm{R}}
\newcommand{\bw}{\bm{w}}
\newcommand{\bu}{\bm{u}}
\newcommand{\bQ}{\bm{Q}}
\newcommand{\bT}{\bm{T}}
\newcommand{\btheta}{\bm{\theta}}

\newcommand{\bK}{\bm{K}}
\newcommand{\bH}{\bm{H}}

\newcommand{\bP}{\bm{P}}
\newcommand{\bp}{\bm{p}}

\newcommand{\Z}{\mathcal{Z}}
\newcommand{\N}{\mathcal{N}}
\newcommand{\X}{\mathcal{X}}

\newcommand{\mtx}[1]{\begin{bmatrix} #1 \end{bmatrix}}
\newcommand{\tr}{\mathsf{T}}

%% file: sections/abstract.tex
\begin{abstract}

    We present a novel Simultaneous Localization and Mapping (SLAM) method that employs Gaussian Process (GP) based landmark (object) representations.
    Instead of conventional grid maps or point cloud registration, we model the environment on a per object basis using GP based contour representations.
    These contours are updated online through a recursive scheme, enabling efficient memory usage.
    The SLAM problem is formulated within a fully Bayesian framework, allowing joint inference over the robot pose and object based map.
    This representation provides semantic information such as the number of objects and their areas, while also supporting probabilistic measurement to object associations.
    Furthermore, the GP based contours yield confidence bounds on object shapes, offering valuable information for downstream tasks like safe navigation and exploration.
    We validate our method on synthetic and real world experiments, and show that it delivers accurate localization and mapping performance across diverse structured environments.

\end{abstract}


%% file: sections/intro.tex
\section{Introduction}
%
%
%
%


\IEEEPARstart{I}{n} mobile robotics, ability to perform simultaneous localization and mapping (SLAM) is a crucial component for smart autonomous robots.
Over the last decades, SLAM for autonomous systems has become a very active research field in engineering.
As a result, numerous real time SLAM algorithms\cite{survey} have emerged using different sensor modalities such as vision based sensors\cite{vbslam} or laser range finders (LiDAR) \cite{liosam, loosely, loosely2}.
Vision based sensors such as monocular or stereo cameras offer rich semantic information about the environment, which is useful for long term mapping and place recognition.
However, these sensors are usually very sensitive to lighting conditions, which affects data quality.
LiDAR sensors offer better robustness against such environmental conditions and have become a very popular sensor choice for mobile robotics.
Recent advancements in sensor technology have made LiDAR sensors lighter, more accurate, and less power hungry while increasing their resolution \cite{mems_lidar}.
As such, these sensors have started to be utilized extensively in mobile robotics in recent works \cite{liosam,iglio,dlio, loam, kiss_icp}.

In this paper, we utilize LiDAR based sensing in our system, focusing on a novel landmark (object) based map representation with statistical confidence bounds.

\begin{figure}[tbp]
    \centering
    \includegraphics[width = \linewidth]{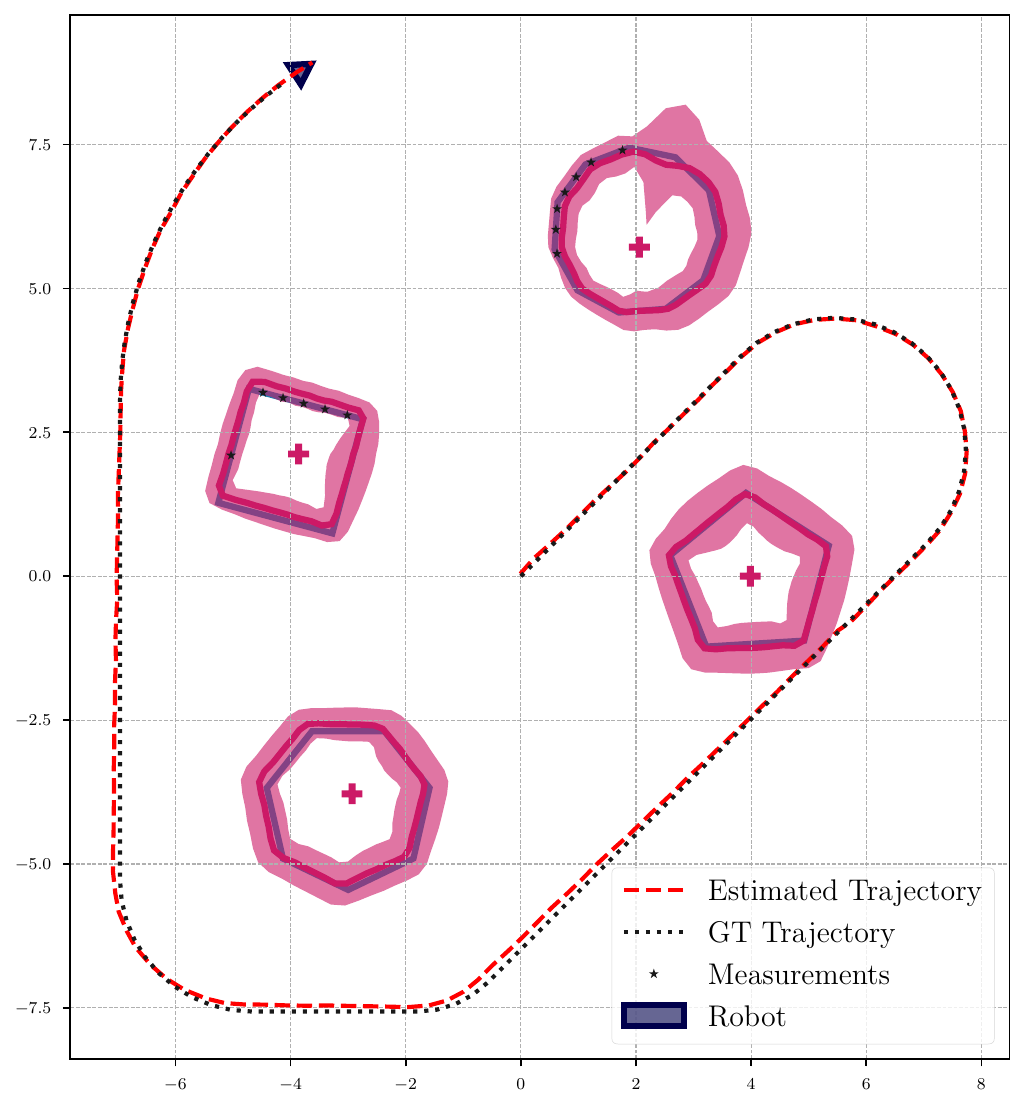}
    \caption{
        Sample output of the proposed method illustrating contour estimates (pink contours), their 99\% confidence regions (pink shaded areas), and the robot trajectory (red dashed line). The method provides mean and covariance estimates of detected objects, which are beneficial for downstream tasks such as planning and exploration.
    }
    \label{fig:visual_abstract}
\end{figure}

\begin{figure*}[tbp]
    \centering
    \includegraphics[width=\linewidth]{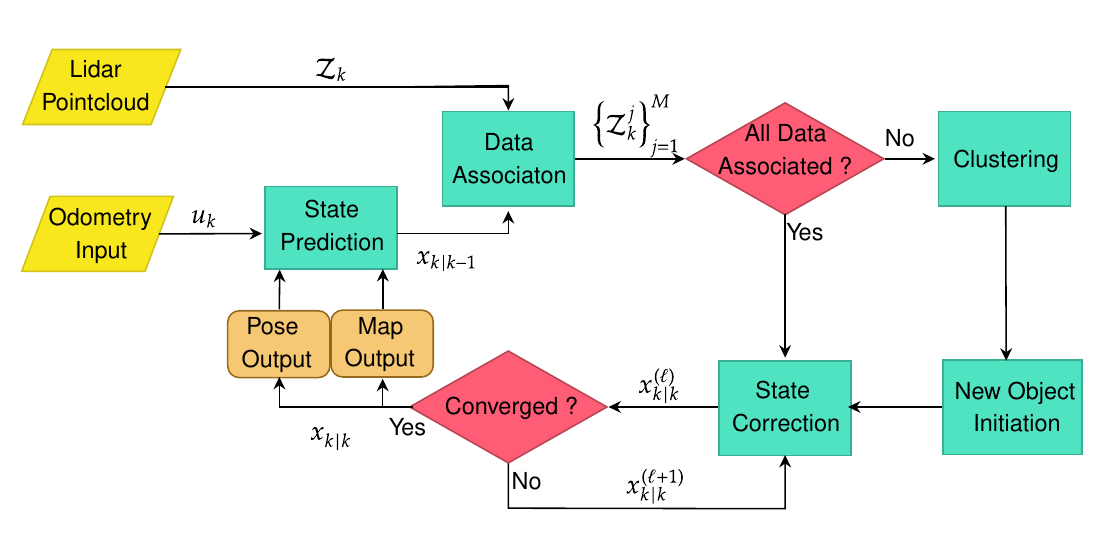}
    \caption{
        The workflow diagram of the proposed system.
        Sensor inputs considered in this setup are 2D point clouds and odometry measurements.
        The system performs clustering, association and object initiation for preprocessing.
        The overall state vector is iteratively corrected until convergence.
        The final map and pose estimates are the system outputs, which are fed back to the system to process the next set of odometry and point cloud inputs.
    }
    \label{fig:diagram}
\end{figure*}

In principle, geometry based LiDAR SLAM methods use geometric entities such as points, lines, and surfaces to establish correspondences between two frames and compute the relative transformation.
Methods like iterative closest point (ICP) \cite{icp}, Generalized ICP (GICP) \cite{gicp}, and their variants are commonly used in the literature to find point to point or point to plane correspondences \cite{iglio, liosam, dlio}.
These methods exhaustively search for correspondences when processing point clouds, leading to high computational complexity that scales with the size of the point cloud.

Earlier approaches, such as \cite{hectorslam, gmapping, gmapping_sonar}, focused on building 2D grid based map representations using filtering based SLAM techniques.
Instead of performing an exhaustive search over all point pairs, feature based methods \cite{feature1, loam, liosam, surfel} use a reduced set of geometric features such as edges, corners, and planes to establish correspondences between two frames, offering a more efficient solution.
However, this efficiency comes at the cost of accuracy, as many potential correspondences are discarded during feature extraction.

In contrast, dense matching based methods \cite{dlio, dense2} utilize the entire point cloud to establish correspondences, resulting in more accurate solutions.
However, to mitigate the high computational cost, these methods often employ a coarser resolution for map representations.

LiDAR based SLAM and odometry techniques can be further categorized according to the manner in which additional sensor data is incorporated.
In loosely coupled methods \cite{loosely, loosely2}, LiDAR measurements are processed independently of other sensor data, which are more robust solutions against conflicting sensor information.
However, because only a subset of the sensor data is used during localization, both the localization accuracy and map accuracy are reduced.
In tightly coupled methods \cite{tight, tight2}, LiDAR measurements are integrated with other sensor information (such as IMU or GPS) for both localization and mapping.

A common feature of these methods is the use of voxel or grid based map representations, which scale poorly with the increasing size of the environment, regardless of object density.
Another drawback of voxel based representations is that they lack any notion of objects, meaning they cannot provide basic information such as the number of objects in the environment or their spatial properties, like area or extent.
Furthermore, obtaining a confidence bound for the map representation is either not possible or computationally expensive.
A statistical representation that provides confidence bounds for object extents would be a valuable addition for downstream tasks such as safe and robust navigation \cite{tomlin, gpis} or exploration \cite{hedac}.

An alternative to voxel based representations is object based map representations.
The vision based SLAM community has proposed several methods for object level representations across various SLAM and localization tasks.
For example, in \cite{cubeslam}, the authors proposed using bounding cube based object representations estimated from 2D bounding boxes.
This framework can handle both static and dynamic objects in the environment.
In \cite{quadricslam}, objects are represented as 3D ellipsoids. The system relies on 2D object detections and generates a factor graph based optimization problem, where the ellipsoids and their positions are updated jointly.

In a series of recent papers \cite{chinese_implicit, chinese_fg}, the authors investigated the use of implicit functions to generate object boundary representations with 2D LiDAR sensors.
In their work, an object is represented by its center point and radial contour function by employing truncated Fourier series representation.
LiDAR measurements are then used to construct a submap of the environment, where the measurements are modeled as factors in a factor graph.
After optimization, Fourier coefficients are updated, and both map representation and robot states are estimated. Although Fourier series based representations are powerful, they don't posses a well defined probability distribution in the spatial domain since the unknown function is represented in the frequency domain. Consequently, the Fourier series based representation is unable to incorporate local updates (e.g., partial measurements from the front or back side of an object); instead, the entire object boundary is updated.


In this context, we introduce GPL-SLAM, a novel Bayesian SLAM algorithm that uses odometry measurements and point cloud data from a LiDAR sensor to estimate robot states and construct a consistent, analytical, and memory efficient map representation.
Unlike traditional voxel or grid based methods, our approach models the environment using an object based representation built on Gaussian process based closed contour descriptions \cite{baum_rhm, ETTGP}.
We employ a finite dimensional approximation of the Gaussian process, enabling us to derive a computationally efficient recursive Bayesian filter for the SLAM problem.
By utilizing Gaussian processes to model the shape function, our algorithm provides a well defined posterior distribution for unknown object boundaries. These distributions provide confidence bounds for object extents, which can also be used to resolve unknown measurement associations. Moreover, we define the Gaussian process model in the spatial domain resulting in a representation that can learn locally.


The effectiveness of our approach is demonstrated in both simulated and real world environments.
Moreover, the statistical properties of our representation make it well suited for downstream applications, such as safe navigation \cite{tomlin} and exploration \cite{hedac}.

%% file: sections/method.tex
\section{Method} \label{ch:method}
This section provides an overview of our proposed system.
We provide the formulation of the SLAM problem and provide a solution using Gaussian Process based object representation.
\subsection{Overview}
Our SLAM algorithm is a robust 2D LiDAR SLAM algorithm that generates object based analytical and compact map representations.
The generated map consists of a set of objects that are modeled by ``star-convex''
object representations, which are commonly used for tracking irregular shaped objects \cite{ETTGP,vbgp, gpett_multi,baum_rhm, bspline, rongli}.
We model the boundary of the star-convex object as a Gaussian Process (GP) and use recursive Bayesian estimation to jointly estimate robot pose and object shapes.

Our system takes a 2D LiDAR pointcloud and an odometry input as system inputs.
The odometry input is used for state prediction while the pointcloud input used for data association first.
If there are any unassociated measurements left, these measurements are clustered and new map objects are initiated for each cluster.
The combined state vector is iteratively updated until convergence using an iterated Kalman filter (IEKF) algorithm \cite{havlik2015performance}.
Map and pose estimates are extracted from the updated state vector as system outputs.
These outputs are fed back to the system to be used in the next time step in a recursive Bayesian manner.
A flowchart of our architecture is depicted on Figure \ref{fig:diagram}.

First, in Section \ref{subsec:formulation} we introduce the SLAM formulation of the proposed approach.
Then, GP based contour formulations are formalized in Section \ref{subsec:contour_model}.
An augmented state space representation of proposed approach is derived in Section \ref{subsec:state_space}.
Finally, proposed data association algorithm is presented in Section \ref{subsec:data_assoc}.

\subsection{SLAM Formulation} \label{subsec:formulation}
In this subsection we introduce the basic formulations and assumptions of our system.
We consider a SLAM scenario on the 2D plane where there are distinct objects to be mapped.
For this task we use a robot equipped with a laser range finder sensor, which can generate measurements from the boundaries of the visible objects.
We denote a set of 2D Cartesian measurements that are obtained at \(k\)-th time step as
\begin{align}
    \Z_k \triangleq \{ \bz_k^j \}_{j=1}^{m_k},
\end{align}
where \(\bz_k^j \in \R^2 \) denotes the \(j\)-th measurement and \(m_k\) denotes the number of measurements at time \(k\).

The robot pose consists of a position vector \(\bx_k^{r,p} \in \R^2\) and an orientation variable \(\theta \in [0, 2\pi]\)
\begin{align} \label{eq:rob_pose_defn}
    \bx_k^r \triangleq \mtx{(\bx_k^{r,p})^\tr & \phi_k }^\tr.
\end{align}
The robot receives \(\Z_k\) from the visible map objects and an odometry input, \(\bu_k \in \R^3\), at each time step.
Various sensors can be used as the odometry source such as IMU, GPS or wheel odometry.

The map objects are tuples consisting of an \textit{object center} and an \textit{object shape function}, which will be detailed in Section \ref{subsec:contour_model}.
We denote map objects as
\begin{align} \label{eq:object_defn}
    \X_k^i \triangleq (\bx_k^{i, c}, f_k^i(\theta)),
\end{align}
where \(\X_k^i\) represents the \(i\)-th object on the map, \(\bx_k^{i,c} \in \R^2\) is the object center poisiton vector and \(f_k^i(\theta)\) is the object shape function.

Our aim is to jointly estimate the augmented state vector using the odometry measurements and 2D point cloud inputs.
To this end, we use recursive Bayesian estimation techniques to estimate the overall augmented state in real time.
In the next subsections, we define our object based representation followed by Gaussian Process based shape estimation framework and state state representation of our full system.

\begin{figure}[tbp]
    \centering
    \begin{subfigure}{.9\linewidth}
        \includegraphics[width=\linewidth]{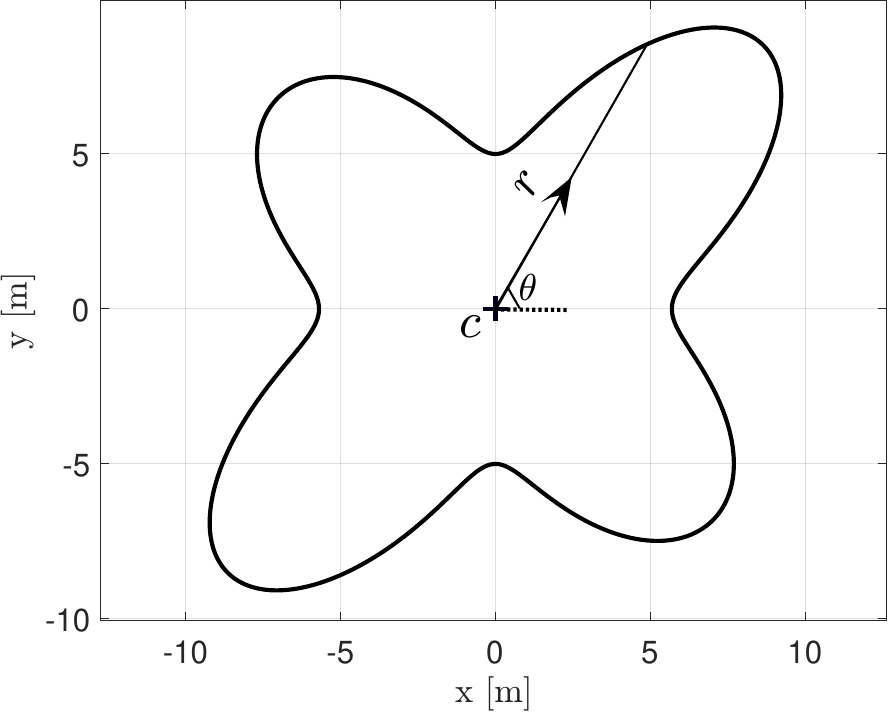}
        \caption{}
        \label{fig:subfigA}
    \end{subfigure}
    \\
    \vspace{1em}
    \begin{subfigure}{.9\linewidth}
        \includegraphics[width=\linewidth]{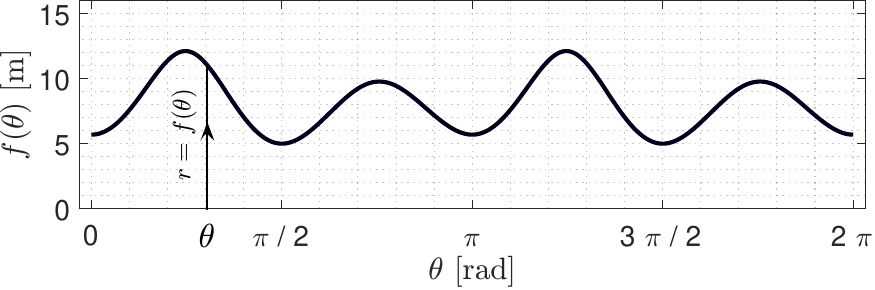}
        \caption{}
        \label{fig:subfigB}
    \end{subfigure}
    \caption{A star-convex shape with its radial representation \(r = f(\theta)\).
        (a) The star-convex shape and its center.
        (b) Radial function representing the star-convex shape parametrization with respect to a center point.}
    \label{fig:radial_representation}
\end{figure}

\subsection{GP Based Object Contour Model} \label{subsec:contour_model}
\subsubsection{Star-Convex Object Model}
Our goal is to efficiently map the objects in the environment by fitting a nonparametric model to the map objects, rather than generating an occupancy grid over the environment or fitting parametric models to object shapes.
To this end, we utilize the \emph{star-convex} shape representation for the objects.
A set \(\mathcal{S}(c)\) is star-convex if the line segment between a fixed center point \(c \in \mathcal{S}\) and any point \(s \in \mathcal{S}\) on the boundary is contained entirely within the set.
Star-convex shapes are frequently used for representing objects in the context of extended object tracking, providing a flexible tool for representing vastly different target geometries \cite{ETTGP,vbgp,baum_rhm,bspline}.
Throughout this paper, we assume that objects in the environment can be represented as star-convex sets.

Boundary of a star-convex set is represented as a polar function \(r = f(\theta)\) parameterized around a center point, as shown in Figure \ref{fig:radial_representation}.
With this representation, the boundary of \(\mathcal{X}^i\), denoted as \(\mathcal{B}^j_\mathcal{G}\) can be parameterized with respect to the global frame as
\begin{align}
    \mathcal{B}^i_\mathcal{G} \triangleq \bigl \{ \bm{b}(\theta) \mid \bm{b}(\theta) = \bm{x}^{i,c} + f^i(\theta) \, \bm{p}(\theta), \ \theta \in [0, 2\pi] \bigr\},
\end{align}
where \(\bm{p}(\theta) \triangleq \mtx{cos(\theta) & \sin(\theta) }^\tr\).
In the global coordinates, we model the measurements generated from the object boundary as noisy samples drawn from \(\mathcal{B}^j_G\)
\begin{align} \label{eq:global_meas}
    \bm{z}^j_\mathcal{G} = \bm{x}^{i,c} + f^i(\theta)\bm{p}(\theta) + \bm{v}, \quad \bm{v} \sim \mathcal{N}(\bm{0}, \bm{R}).
\end{align}
Consequently, measurements in the local (robot) frame can be written using a coordinate transform
\begin{align} \label{eq:local_meas}
    \bm{z}^j_\mathcal{L} = \bm{T}^\tr (\bx^{i,c} + f^i(\theta)\bp(\theta) - \bm{x}^{r,p}) + \bv, \quad \N(\bm{0}, \bR) ,
\end{align}
where the rotation matrix is defined as
\begin{align}
    \bT \triangleq \mtx{\cos(\phi) & -\sin(\phi) \\ \sin(\phi) & \cos(\phi)},
\end{align}
and \(\phi\) is orientation angle of the robot.

With this setup, the robot observes noisy measurements of \(f(\theta)\) in the form of 2D points.
With each measurement, more information about \(f(\theta)\) is obtained, and this information can be used to estimate the radial function associated with an object.
To estimate the radial function using noisy measurements, we employ a Gaussian Process (GP) regression method.

\subsubsection{Gaussian Processes}
We utilize Gaussian Processes to model the radius function $f(\theta)$.
Gaussian processes are infinite dimensional generalizations of the multivariate Gaussian variables over functions \cite[Ch. 6.4]{bishop}.
A Gaussian process $f(u)$ is uniquely determined by its mean and covariance functions
\begin{subequations}\label{eqn:gp_form}
    \begin{align}
        m(u)                        & =\mathbb{E}[f(u)], \label{gp_form:a}                                                                   \\
        k\left(u, u^{\prime}\right) & =\mathbb{E}\left[(f(u)-\mu(u))\left(f\left(u^{\prime}\right)-\mu\left(u^{\prime}\right)\right)\right],
    \end{align}
\end{subequations}

and $f(u)$ is denoted as:

\begin{equation}
    f(u) \sim  \mathcal{GP}\left(m(u), k\left(u, u^{\prime}\right)\right)
\end{equation}

Typically, a function is estimated offline using GP regression methods and then deployed for online inference  \cite{Rasmussen2004, deisenroth1, deisenroth2, zeilinger, tomlin}.
In our use case, we utilize GPs in an online learning context where the robot must update the shape estimate in real time and data gathered from the LiDAR contains many redundant points.
The recursive GP regression technique has been used widely in the context of online target tracking which provides an approximate but fast alternative for full GP formulation \cite{huber,ETTGP,vbgp,3dgpett} .

Consider a vector of function evaluations in the form
\begin{align}
    \bx^f \triangleq \mtx{f(\theta_1) & f(\theta_2) & \dots & f(\theta_N)}^\tr,
\end{align}
where \(\btheta \triangleq \mtx{ \theta_1 & \theta_2 & \dots, \theta_N }\) are predefined fixed input points.
For \(\bx^f\) we construct a linear state space formulation in order to apply a Kalman filter.
For the recursive GP formulation we consider a linear measurement model
\begin{align}
    y_k = \bH^f_k(\theta_k') \bx^f_k + e^f_k(\theta_k'), \quad e^f_k(\theta_k') \sim \N(0, R^f_k(\theta_k'))
\end{align}
where \(y_k\) is the \(k\)th value obtained from the function at the evaluation point \(\theta_k'\), \(\bH^f_k(\theta_k') \) is a measurement matrix and \(e^f_k(\theta_k')\) is the measurement noise term.
As in \cite{ETTGP}, equations that determine \(\bH^f_k(\theta_k')\) and \(e_k^f(\theta_k')\) are evaluated as
\begin{subequations} \label{eq:recursive_gp_hr}
    \begin{align}
        \bH_k^f (\theta_k') & = \bK(\theta_k', \btheta) (\bK(\btheta, \btheta))^{-1},                                                        \\
        R_k^f (\theta_k')   & = k(\theta_k', \theta_k') + R - \bK(\theta_k', \btheta)  (\bK(\btheta, \btheta))^{-1} \bK(\btheta, \theta_k'),
    \end{align}
\end{subequations}
where \(R\) is a fixed measurement noise term and
\begin{align}
    \bK(\bm{\alpha}, \bm{\beta}) \triangleq \mtx{
    k(\alpha_1, \beta_1) & \dots & k(\alpha_1, \beta_I) \\
    \vdots               & \     & \vdots               \\
    k(\alpha_J, \beta_1) & \dots & k(\alpha_J, \beta_I) \\
    }
\end{align}
for \(\bm{\alpha} \in \R^I\), \(\bm{\beta} \in \R^J\).

Assuming a constant function is being estimated, as state space formulation for the radial function is formulated as

\begin{subequations} \label{eq:extent_state_space}
    \begin{align}
        \begin{split} \label{eq:recursive_gp_motion}
            \bx_{k+1}^f & = \bx_k^f + \bw_k^f, \quad \bw_k^f \sim \N(0, \bQ_k^f),
        \end{split}               \\
        \begin{split} \label{eq:recursive_gp_meas}
            y_k & = \bH_k^f(\theta_k) \bx_k^f + e_k^f, \quad e_k^f \sim \N(0, R_k^f(\theta_k)),
        \end{split} \\
        \begin{split} \label{eq:recursive_gp_init}
            \bx_0^f & \sim \N(\bm{0}, \bP_0^f),
        \end{split}
    \end{align}
\end{subequations}
where \(\bw_k^f\) is a user defined process noise term and \(\bP_0^f \triangleq \bK(\btheta, \btheta)\).

\subsubsection{Mean Function} \label{subsec:mean_func}
We consider a constant but unknown mean function \(m(\theta) = r\), where \(r\) represents the mean radius of the object \cite{ETTGP}.
The associated Gaussian Process (GP) is defined as
\begin{align} \label{eq:mean_func}
    f(\theta) \sim \mathcal{GP}(r, k(\theta, \theta')), \quad r \sim \mathcal{N}(0, \sigma_r^2).
\end{align}
This setup is equivalent to a zero-mean GP but with an additional variance term
\begin{align} \label{eq:mean_func_cov_equivalent}
    f(\theta) \sim \mathcal{GP}(0, k(\theta, \theta') + \sigma_r^2).
\end{align}

\subsubsection{Covariance Function} \label{subsec:cov_func}
The covariance function in GP modeling plays a crucial role in capturing the underlying spatial or temporal dependencies in the data.
It quantifies the similarity between input points and dictates how much influence nearby points have on each other's function evaluations within the GP framework.
One of the most commonly used covariance functions is the squared exponential kernel
\begin{align} \label{eq:se_kernel}
    k(\theta, \theta') = \sigma_f^2 \exp \biggl(-{\frac{2(\theta - \theta')^2}{l^2}} \biggr),
\end{align}
where \(\sigma_f^2\) represents the prior variance of the function amplitude and \(l\) is referred to as the length scale.
The length scale quantifies how quickly the covariance decays with distance between points \(\theta\) and \(\theta'\).

For our purposes, a periodic kernel is more desirable since we want to model object shape functions as periodic radial functions with periodicity \(2\pi\).
In this setting, \(f(\theta)\) and \(f(\theta + 2\pi)\) should be perfectly correlated, i.e
\begin{align}
    k(\theta, \theta+2\pi) = k(\theta, \theta).
\end{align}
To enforce periodicity, \eqref{eq:se_kernel} can be modified as
\begin{align} \label{eq:pse_kernel}
    k(\theta, \theta') = \sigma_f^2 \exp \biggl({-\frac{2\sin(|\theta - \theta'|)^2}{l^2}} \biggr),
\end{align}
which is also referred to as the periodic kernel \cite{ETTGP}.
The final form of our covariance function will include a combination of the periodic kernel with the mean function related radial variance term introduced in Section \ref{subsec:mean_func}.
After combining \eqref{eq:pse_kernel} and \eqref{eq:mean_func_cov_equivalent} the final form of the kernel is
\begin{align} \label{eq:my_kernel}
    \bar{k}(\theta, \theta') \triangleq k(\theta, \theta') + \sigma_r^2 = \sigma_f^2 \exp \biggl({-\frac{2\sin(|\theta - \theta'|)^2}{l^2}} \biggr) + \sigma_r^2.
\end{align}

\subsection{State Space Representation} \label{subsec:state_space}
We consider an augmented state vector to formulate recursive inference equations in a compact way.
The state vector consists of the robot pose \(\bx_k^r\), object centers \(\{\bx_k^{i,c}\}_{i=1}^M\) and and object shapes \(\{\bx_k^{i,f}\}_{i=1}^M\)
\begin{align}
    \bx_k \triangleq \mtx{
    (\bx_k^r)^\tr & (\bx_k^{1,c})^\tr & (\bx_k^{1,f})^\tr & \dots & (\bx_k^{M,c})^\tr & (\bx_k^{M,f})^\tr
    }^\tr,
\end{align}
where \(\bx_k^r\) is defined as in \eqref{eq:rob_pose_defn}, \(\{\bx_k^{i,c}\}_{i=1}^M\) are defined as in \eqref{eq:object_defn} and \(\{\bx_k^{i,f}\}_{i=1}^M\) are defined as the vectors corresponding to the object contour functions in \ref{subsec:contour_model}.

\subsubsection{Measurement Model}
\begin{figure}[tbp]
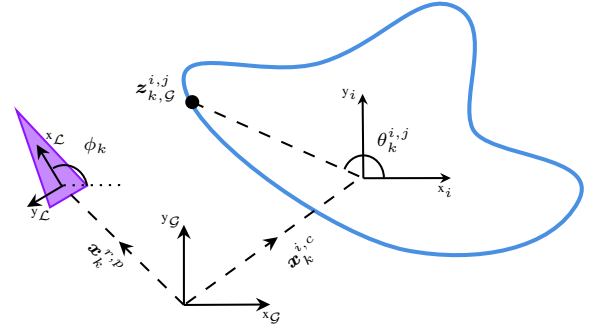

    \centering
    \includestandalone[width=.9\linewidth]{figures/example_setup}
    \caption{
        A visualization of the robot pose and \(j\)th object at time \(k\).
        A measurement is generated from the object contour, \(\bz_{k,\mathcal{G}}^{i,j}\), at angle \(\theta_k^{i,j}\) relative to the object center \(\bx_k^{i,c}\).
        The robot position \(\bx_k^{r,p}\) and orientation \(\phi_k\) determines the local coordinate frame of the robot.
    }
    \label{fig:example_setup}
\end{figure}

Assuming known correspondences, i.e. the perfect knowledge as to which measurement belongs to which object, we denote local measurement \(\bz_{k, \mathcal{L}}^{i,j}\) as the \(j\)th measurement associated with the \(i\)th object.
The angular location of the measurement is determined with respect to center position of the landmark, as depicted in Figure \ref{fig:example_setup}, as
\begin{align}
    \theta_k^{i,j} \triangleq \angle(\bT_k\bz_{k, \mathcal{L}}^{i,j} + \bx_k^{r,p} - \bx_k^{i, c}),
\end{align}
which can be used to find the unit direction vector from the object center to the measurement
\begin{align}
    \bp( \theta_k^{i,j}) = \mtx{\cos(\theta_k^{i,j}) \\ \sin(\theta_k^{i,j})} = \frac{\bT_k\bz_{k, \mathcal{L}}^{i,j} + \bx_k^{r,p} - \bx_k^{i, c}}{||\bT_k\bz_{k, \mathcal{L}}^{i,j} + \bx_k^{r,p} - \bx_k^{i, c}||}.
\end{align}

Combining \eqref{eq:local_meas} and \eqref{eq:recursive_gp_meas}, we obtain
\begin{align}
    \bz_{k, \mathcal{L}}^{i,j} & = \bT_k^\tr(\bx_k^{i, c} + \bp(\theta_k^{i,j})  (\bH_k^f(\theta_k^{i,j}) \bx_k^{i,f} + e_k^f)) + \bv_k^{i,j}, \nonumber \\
                               & = \bT_k^\tr (\bx_k^{i, c} + \bp(\theta_k^{i,j}) \bH_k^f(\theta_k^{i,j}) \bx_k^{i,f}  - \bx_k^{r,p}) \nonumber           \\
                               & + \bT_k^\tr \bp(\theta_k^{i,j}) e_k^f + \bv_k^{i,j},
\end{align}
which can be written in a compact form
\begin{align} \label{eq:single_meas_model}
    \bz_{k, \mathcal{L}}^{i,j} & = h_k^{i, j}(\bx_k) + \bar{\bv}_k^{i,j}, \quad \bar{\bv}_k^{i,j} \sim \N(\bm{0}, \bR_k^{i,j})
\end{align}
where
\begin{subequations}
    \begin{align}
        h_k^{i,j}(\bx_k)  & \triangleq \bT_k^\tr (\bx_k^{i, c} + \bp(\theta_k^{i,j}) \bH_k^f(\theta_k^{i,j}) \bx_k^{i,f} - \bx_k^{r,p}), \\
        \bar{\bv}_k^{i,j} & \triangleq \bT_k^\tr (\bp(\theta_k^{i,j}) e_k^f) + \bv_k^{i,j},                                              \\
        \bR_k^{i,j}       & =\bT_k^\tr \bp(\theta_k^{i,j}) R_k^f(\theta_k^{i,j})\bp(\theta_k^{i,j})^\tr \bT_k + \bR
    \end{align}
\end{subequations}
where \(h_k^{i,j}(.)\) and \(\bv_k^{i,j}\) can be interpreted as measurement functions and noise model related terms, respectively.

\subsubsection{Motion Model}

The robot pose and map objects evolve in time in different ways.
We consider the state space equations
\begin{subequations}\label{eq:individual_motion_models}
    \begin{align}
        \bx_{k+1}^r     & = \bx_k^r + \bu_k^r + \bw_k^r, \quad & \bw_k^r \sim \N(\bm{0}, \bQ^r),        \\
        \bx_{k+1}^{i,c} & = \bx_k^{i,c} + \bw_k^c, \quad       & \bw_k^c       \sim \N(\bm{0}, \bQ^c),  \\
        \bx_{k+1}^{i,f} & = \bx_k^{i,f} + \bw_k^f \quad        & \bw_k^f        \sim \N(\bm{0}, \bQ^f),
    \end{align}
\end{subequations}
where \(\bQ^r\) and \(\bQ^c\) are covariance matrices and \(\bQ^f\) is chosen according to the kernel function to preserve periodicity of the radial shape function.


The overall motion model can be obtained as an augmented form of the equations in \eqref{eq:individual_motion_models}
\begin{align} \label{eq:full_motion_model}
    \bx_{k+1} = \bx_k + \bu_k + \bw_k, \quad \bw_k \sim \N(\bm{0}, \bQ),
\end{align}
where
\begin{subequations}
    \begin{align}
        \bu_k & \triangleq \mtx{(\bu_k^r)^\tr                                                    & \bm{0} & \dots & \bm{0} }^\tr, \\
        \bQ   & \triangleq \text{blkdiag}(\bQ^r, \, \bQ^c, \, \bQ^f, \, \dots, \bQ^c, \, \bQ^f).
    \end{align}
\end{subequations}

\subsubsection{Inference}

A sequential update rule is essential for online SLAM.
Using the Bayesian estimation framework, an IEKF based solution to the SLAM problem can be obtained using full state space equations.
To apply the IEKF formulation, we augment the measurements obtained from each object, \(\{\bz_k^{i,j}\}_{j=1}^{m_k^i}\), and their corresponding measurement models as
\begin{subequations} \label{eq:augment_meas_per_object}
    \begin{align}
        \bz_k^i       & \triangleq \mtx{(\bz_k^{i, 1})^\tr                                              & (\bz_k^{i, 2})^\tr   & \dots & (\bz_k^{i, m_k^i})^\tr          & }^\tr, \\
        h_k^i (\bx_k) & \triangleq \mtx{h_k^{i,1}(\bx_k)^\tr                                            & h_k^{i,2}(\bx_k)^\tr & \dots & h_k^{i, m_k^i}(\bx_k)^\tr}^\tr,          \\
        \bR_k^i       & \triangleq \text{blkdiag}(\bR_k^{i,1}, \bR_k^{i,2},  \dots,  \bR_k^{i, m_k^i}).
    \end{align}
\end{subequations}
Using the augmented model variables, a measurement model can be constructed for all measurements obtained at time \(k\)
\begin{align} \label{eq:full_meas_model}
    \bz_k = h_k(\bx_k) + \bv_k, \quad \bv_k \sim \N(0, \bR_k),
\end{align}
where
\begin{subequations}
    \begin{align}
        \bz_k      & \triangleq \mtx{(\bz_k^1)^\tr                                                & (\bz_k^2)^\tr & \dots & (\bz_k^M)^\tr & }^\tr, \\
        h_k(\bx_k) & \triangleq \text{blkdiag}(h_k^1(\bx_k),  h_k^2(\bx_k), \dots, h_k^M(\bx_k)),                                                  \\
        \bR_k      & \triangleq \text{blkdiag}(\bR_k^1, \bR_k^2,  \dots,  \bR_k^M).
    \end{align}
\end{subequations}
The final form of the state space equations is
\begin{subequations}\label{eq:augmented_full_state_space}
    \begin{align}
        \bx_{k+1} & = \bx_k + \bu_k + \bw_k, \quad & \bw_k \sim \N(\bm{0}, \bQ),   \\
        \bz_k     & = h_k(\bx_k) + \bv_k, \quad    & \bv_k \sim \N(\bm{0}, \bR_k), \\
        \bx_0     & \sim \N(\hat{\bx}_0, \bSig_0), & \
    \end{align}
\end{subequations}
where \(\hat{\bx}_0\) and \(\bSig_0\) are the initial mean and covariance variables of the overall system.
The full state space formulation can be used to perform online SLAM with the help of an iterated extended Kalman filter.

\subsection{Data Association} \label{subsec:data_assoc}

To address the data association problem, we exploit the GP representation of the object contour function. Specifically, we exploit the uncertainty bounds given by the GP representing the radial function to perform likelihood based data association as well as outlier rejection.
Given the GP representation of an object contour \(\bx_k^{i,f}\) and the input point \(\theta_k^{i,j}\) from which a measurement \(\bz_k^j\) is obtained, we utilize the likelihood function
\begin{align}
    \ell_k^{i,j} \triangleq p(f^i(\theta_k^{i,j}) = r \mid \Z_{1:k-1}) = \N(r; \mu^{i,j}_k, (\sigma^{i,j}_k)^2),
\end{align}
where \(r\) is defined as
\begin{align}
    r \triangleq || \bT_k \bz_{k, \mathcal{L}}^j + \bx_k^{r,p} - \bx_k^{i,c}||,
\end{align}
to facilitate the data association process. This likelihood serves as a key component in determining the most probable correspondence between measurements and objects.

The sufficient statistics at the input point \(\theta_k^{i,j}\) can be found using the current object estimates and the prior measurements.
Specifically, the marginal distribution of the GP at \(\theta_k^{i,j}\) given the previous measurements is computed as
\begin{align} \label{eq:marginal_extent_gp}
    p(f^i(\theta_k^{i,j}) \mid \Z_{1:k-1}) & = \int p(f^i(\theta_k^{i,j}) \mid \bx_k^{i,f}) \nonumber \\
                                           & \times p(\bx_k^{i,f} \mid \Z_{1:k-1}) \, d\bx_k^{i,f},
\end{align}
which leads to the mean and variance expressions  \cite[Ch. 15]{murphy2013machine}
\begin{subequations}
    \begin{align}
        \mu_k^{i,j}        & \triangleq \bK(\theta_k^{i,j}, \btheta)\bK(\btheta, \btheta)^{-1} \hat{\bx}_{k \mid k-1}^{i,f},                                                           \\
        (\sigma^{i,j}_k)^2 & \triangleq k(\theta_k^{i,j}, \theta_k^{i,j}) + \bK(\theta_k^{i,j}, \btheta)(\bK(\btheta, \btheta)^{-1} \bSig_{k \mid k-1}^{i,f} - \mathbb{I}_N) \nonumber \\
                           & \times\bK(\btheta, \btheta)^{-1}\bK(\btheta, \theta_k^{i,j}).
    \end{align}
\end{subequations}

The likelihood function provides a mechanism for rejecting measurements that are likely to have originated from external sources, thereby reducing the risk of false associations \cite{bar1975tracking}.
This is achieved by utilizing a squared Mahalanobis distance criterion \cite{ghorbani2019mahalanobis}, where measurements are rejected if
\begin{align} \label{eq:gating}
    (r - \mu_k^{i,j}) (\sigma_k^{i,j})^{-2}(r - \mu_k^{i,j}) \geq \gamma.
\end{align}
The left-hand side of Eq. \eqref{eq:gating} follows a chi-square distribution with one degree of freedom. The gate size threshold \(\gamma \in \R_{+}\) can be determined using the quantiles (e.g. $95 \%$) of the chi-square distribution.
This can be interpreted as whenever the distance exceeds a certain threshold, the likelihood becomes zero, i.e. the measurement is rejected.
This approach ensures that only measurements with a high likelihood of accurate association are considered, thus enhancing the overall robustness of the data association process.
Additionally, it can be employed to identify the most likely candidate for association, which is determined by
\begin{align} \label{eq:max_lkhood_assoc}
    \hat{i} = \arg \max_i \ell_k^{i,j}.
\end{align}

Note that the proposed association strategy is generic in the sense that it does not prohibit the use of existing data association and clustering methods. For instance, measurements can initially be grouped using standard clustering methods, after which the proposed likelihood or predicted contour can be employed to associate clusters with the most probable object.

Furthermore, the predicted contour representation can be exploited in a point-to-point matching procedure similar to those found in ICP and its variants.
The covariance information, hence the confidence bounds may also be used to weight the contribution of each measurement to the association process.

In addition, this form of data association can be thought of as a \textit{ direct} approach \cite{c_sensor,ergodic_matching} or corresponce-free approach as opposed to the \textit{indirect} approach where the correspondence is first established and then the transformations are computed. This means that the proposed solution does not require storing the past measurements for correspondences while providing robustness to cases where measurements are sparse and correspondence matching is problematic.


\subsection{Summary}
Our approach models the map objects as star-convex sets, contrary to the classical models where the map objects are modeled as points.
We use the GP representation to estimate the radial contour function of the star-convex map objects and utilize a recursive formulation to enable online SLAM.
We formulate the whole problem as a state estimation problem by augmenting the robot and object states in a single state vector.
To enable online SLAM we used an IEKF formulation to obtain robot pose and object contour estimates.
This method essentially combines the flexibility of the GP model and the computational efficiency of the IEKF solution.

\begin{algorithm}[tbp]
    \caption{The Proposed SLAM Algorithm} \label{alg:slam}
    \begin{algorithmic}
        \State \textbf{Input:} \(\Z_{k}\), \(\hat{\bx}_{k-1 \mid k-1}\), \(\bSig_{k-1 \mid k-1}\)

        \State \textbf{Prediction Update:}
        \State Obtain \(\hat{\bx}_{k \mid k-1}\) and \(\bSig_{k \mid k-1}\) using \eqref{eq:full_motion_model}

        \If{\(\Z_k \neq \emptyset\)}
        \State \textbf{Data Association:}
        \State Calculate \(\ell_k^{i,j}\) for all object-measurement tuples \((\bx_k^i, \bz_k^j)\)
        \State Perform measurement rejection using \eqref{eq:gating}
        \State Associate measurements and objects using \eqref{eq:max_lkhood_assoc}
        \State Collect unassociated measurements in a set denoted as \(\Z_k'\)

        \If{\(\Z_k' \neq \emptyset \)}
        \State \textbf{Object Initiation:}
        \State Cluster \(\Z_k'\) using a clustering method of choice \cite{saxena2017review}
        \State Initiate new map objects for each cluster
        \State Append new state means \(\hat{\bx}_{k \mid k-1}'\) to \(\hat{\bx}_{k \mid k-1}\)
        \State Append new covariances \(\bSig_{k \mid k-1}'\) to \(\bSig_{k \mid k-1}\)
        \EndIf

        \State \textbf{Correction Update:}
        \For{\(i \gets 1\) to \(N\)}
        \State Obtain augmented measurement vector \(\bz_k^i\) as in \eqref{eq:augment_meas_per_object}
        \State Obtain augmented measurement function \(h_k^{i}(\bx_k)\) as in \eqref{eq:augment_meas_per_object}
        \State Obtain augmented measurement noise covariance matrix \(\bR_k\) as in \eqref{eq:augment_meas_per_object}
        \EndFor
        \State Obtain augmented measurement model as in \eqref{eq:full_meas_model}
        \State Perform IEKF update until convergence as described in Appendix \ref{ap:iekf}
        \EndIf

        \State \textbf{Output:} \(\hat{\bx}_{k \mid k}\), \(\bSig_{k \mid k}\)
    \end{algorithmic}
\end{algorithm}

%% file: figures/example_setup.tex
\tikzset{every picture/.style={line width=0.75pt}} 

\begin{tikzpicture}[x=0.75pt,y=0.75pt,yscale=-1,xscale=1]

    \draw  [dash pattern={on 4.5pt off 4.5pt}]  (247.61,63.32) -- (331.89,100.82) ;
    \draw  [color={rgb, 255:red, 74; green, 144; blue, 226 }  ,draw opacity=1 ][line width=1.5]  (244.33,57.67) .. controls (236.07,31.34) and (281.82,53) .. (311.67,43.67) .. controls (341.51,34.33) and (359.57,7.16) .. (378.05,16.88) .. controls (396.52,26.61) and (379,61.67) .. (385.67,76.33) .. controls (392.33,91) and (450.28,88.61) .. (437.67,112.33) .. controls (425.05,136.06) and (387.67,143) .. (353,136.33) .. controls (318.33,129.67) and (252.59,83.99) .. (244.33,57.67) -- cycle ;
    \draw  [color={rgb, 255:red, 144; green, 19; blue, 254 }  ,draw opacity=1 ][fill={rgb, 255:red, 144; green, 19; blue, 254 }  ,fill opacity=0.5 ] (161.14,67.49) -- (195.82,104.51) -- (177.57,115.18) -- cycle ;
    \draw [line width=0.75]    (331.89,100.82) -- (372,100.75) ;
    \draw [shift={(375,100.75)}, rotate = 179.91] [fill={rgb, 255:red, 0; green, 0; blue, 0 }  ][line width=0.08]  [draw opacity=0] (5.36,-2.57) -- (0,0) -- (5.36,2.57) -- (3.56,0) -- cycle    ;
    \draw [line width=0.75]    (183.1,104.29) -- (172.55,86.77) ;
    \draw [shift={(171,84.2)}, rotate = 58.94] [fill={rgb, 255:red, 0; green, 0; blue, 0 }  ][line width=0.08]  [draw opacity=0] (7.14,-3.43) -- (0,0) -- (7.14,3.43) -- (4.74,0) -- cycle    ;
    \draw  [draw opacity=0] (322.58,96.41) .. controls (324.14,92.36) and (327.78,89.53) .. (331.99,89.57) .. controls (337.45,89.62) and (341.86,94.46) .. (342.01,100.46) -- (331.89,100.82) -- cycle ; \draw   (322.58,96.41) .. controls (324.14,92.36) and (327.78,89.53) .. (331.99,89.57) .. controls (337.45,89.62) and (341.86,94.46) .. (342.01,100.46) ;
    \draw [line width=0.75]    (184.2,104.31) -- (168.78,113.47) ;
    \draw [shift={(166.2,115)}, rotate = 329.28] [fill={rgb, 255:red, 0; green, 0; blue, 0 }  ][line width=0.08]  [draw opacity=0] (7.14,-3.43) -- (0,0) -- (7.14,3.43) -- (4.74,0) -- cycle    ;
    \draw [line width=0.75]    (331.89,100.82) -- (331.53,62.25) ;
    \draw [shift={(331.5,59.25)}, rotate = 89.46] [fill={rgb, 255:red, 0; green, 0; blue, 0 }  ][line width=0.08]  [draw opacity=0] (5.36,-2.57) -- (0,0) -- (5.36,2.57) -- (3.56,0) -- cycle    ;
    \draw [line width=0.75]    (243.51,163.29) -- (243.5,126.75) ;
    \draw [shift={(243.5,123.75)}, rotate = 89.99] [fill={rgb, 255:red, 0; green, 0; blue, 0 }  ][line width=0.08]  [draw opacity=0] (7.14,-3.43) -- (0,0) -- (7.14,3.43) -- (4.74,0) -- cycle    ;
    \draw [line width=0.75]    (243.51,163.29) -- (283.05,163.09) ;
    \draw [shift={(286.05,163.07)}, rotate = 179.7] [fill={rgb, 255:red, 0; green, 0; blue, 0 }  ][line width=0.08]  [draw opacity=0] (7.14,-3.43) -- (0,0) -- (7.14,3.43) -- (4.74,0) -- cycle    ;
    \draw  [dash pattern={on 0.84pt off 2.51pt}]  (184.2,104.31) -- (212.2,104.2) ;
    \draw  [draw opacity=0] (179.06,95.92) .. controls (180.7,94.92) and (182.59,94.36) .. (184.6,94.36) .. controls (190.2,94.36) and (194.85,98.74) .. (195.82,104.51) -- (184.6,106.76) -- cycle ; \draw   (179.06,95.92) .. controls (180.7,94.92) and (182.59,94.36) .. (184.6,94.36) .. controls (190.2,94.36) and (194.85,98.74) .. (195.82,104.51) ;
    \draw [color={rgb, 255:red, 0; green, 0; blue, 0 }  ,draw opacity=1 ][line width=0.75]  [dash pattern={on 4.5pt off 4.5pt}]  (243.51,163.29) -- (183.1,104.29) ;
    \draw [shift={(211.45,131.97)}, rotate = 44.33] [fill={rgb, 255:red, 0; green, 0; blue, 0 }  ,fill opacity=1 ][line width=0.08]  [draw opacity=0] (7.14,-3.43) -- (0,0) -- (7.14,3.43) -- (4.74,0) -- cycle    ;
    \draw [color={rgb, 255:red, 0; green, 0; blue, 0 }  ,draw opacity=1 ][line width=0.75]  [dash pattern={on 4.5pt off 4.5pt}]  (244.56,162.61) -- (330.84,100.14) ;
    \draw [shift={(289.81,129.85)}, rotate = 144.09] [fill={rgb, 255:red, 0; green, 0; blue, 0 }  ,fill opacity=1 ][line width=0.08]  [draw opacity=0] (7.14,-3.43) -- (0,0) -- (7.14,3.43) -- (4.74,0) -- cycle    ;
    \draw  [fill={rgb, 255:red, 0; green, 0; blue, 0 }  ,fill opacity=1 ] (244.67,63.32) .. controls (244.67,61.69) and (245.98,60.37) .. (247.61,60.37) .. controls (249.24,60.37) and (250.56,61.69) .. (250.56,63.32) .. controls (250.56,64.94) and (249.24,66.26) .. (247.61,66.26) .. controls (245.98,66.26) and (244.67,64.94) .. (244.67,63.32) -- cycle ;

    \draw (218,48.6) node [anchor=north west][inner sep=0.75pt]  [font=\scriptsize]  {$\bz_{k,\mathcal{G}}^{i,j}$};
    \draw (288.97,138.23) node [anchor=north west][inner sep=0.75pt]  [font=\scriptsize,rotate=-322.26]  {$\bx_{k}^{i,c}$};
    \draw (337.03,74.4) node [anchor=north west][inner sep=0.75pt]  [font=\scriptsize]  {$\theta_{k}^{i,j}$};
    \draw (367.1,103.3) node [anchor=north west][inner sep=0.75pt]  [font=\tiny]  {$\text{x}_{i}$};
    \draw (320,55.1) node [anchor=north west][inner sep=0.75pt]  [font=\tiny]  {$\text{y}_{i}$};
    \draw (174,77) node [anchor=north west][inner sep=0.75pt]  [font=\tiny]  {$\text{x}_{\mathcal{L}}$};
    \draw (166.7,114.5) node [anchor=north west][inner sep=0.75pt]  [font=\tiny]  {$\text{y}_{\mathcal{L}}$};
    \draw (201.12,127.89) node [anchor=north west][inner sep=0.75pt]  [font=\scriptsize,rotate=-46.1]  {$\bx_{k}^{r,p}$};
    \draw (280.73,165.14) node [anchor=north west][inner sep=0.75pt]  [font=\tiny]  {$\text{x}_\mathcal{G}$};
    \draw (231.04,117.48) node [anchor=north west][inner sep=0.75pt]  [font=\tiny]  {$\text{y}_\mathcal{G}$};
    \draw (192.51,78.97) node [anchor=north west][inner sep=0.75pt]  [font=\scriptsize]  {$\phi _{k}$};

\end{tikzpicture}

%% file: sections/results.tex
\section{Experiments and Results} \label{ch:results}
In this chapter, we present and analyze the experimental results obtained from the implementation of our proposed approach. We first demonstrate our method in a series of simulation studies consisting of simple geometric shapes and more complex irregular shapes. Next, we present the results of our method in a real data scenario where a custom mobile platform equipped with a LiDAR sensor is used to collect data from a parking lot. Further discussion of the results is provided in Section \ref{sec:discussion}.




\begin{figure*}[tbp]
    \centering
    \includegraphics[width=.32\textwidth]{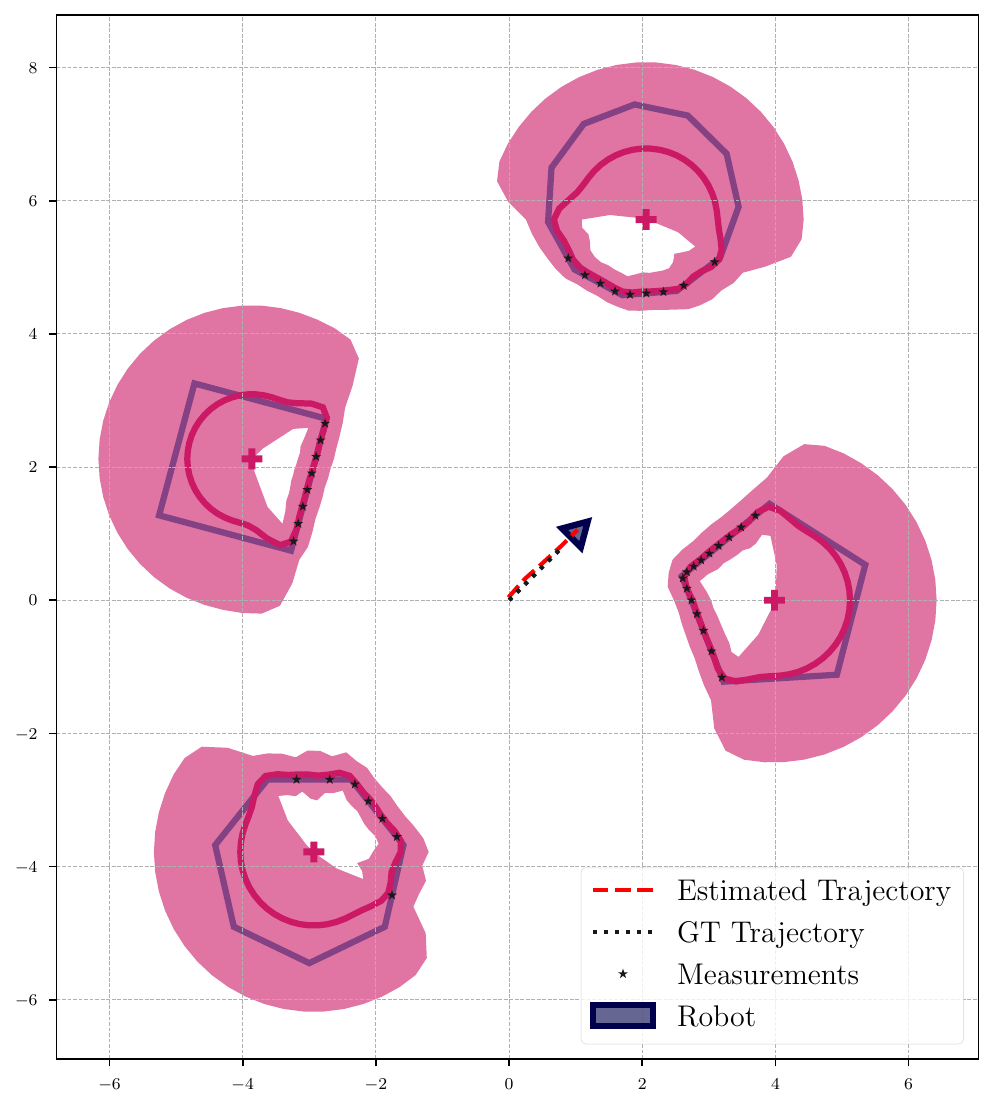}
    \includegraphics[width=.32\textwidth]{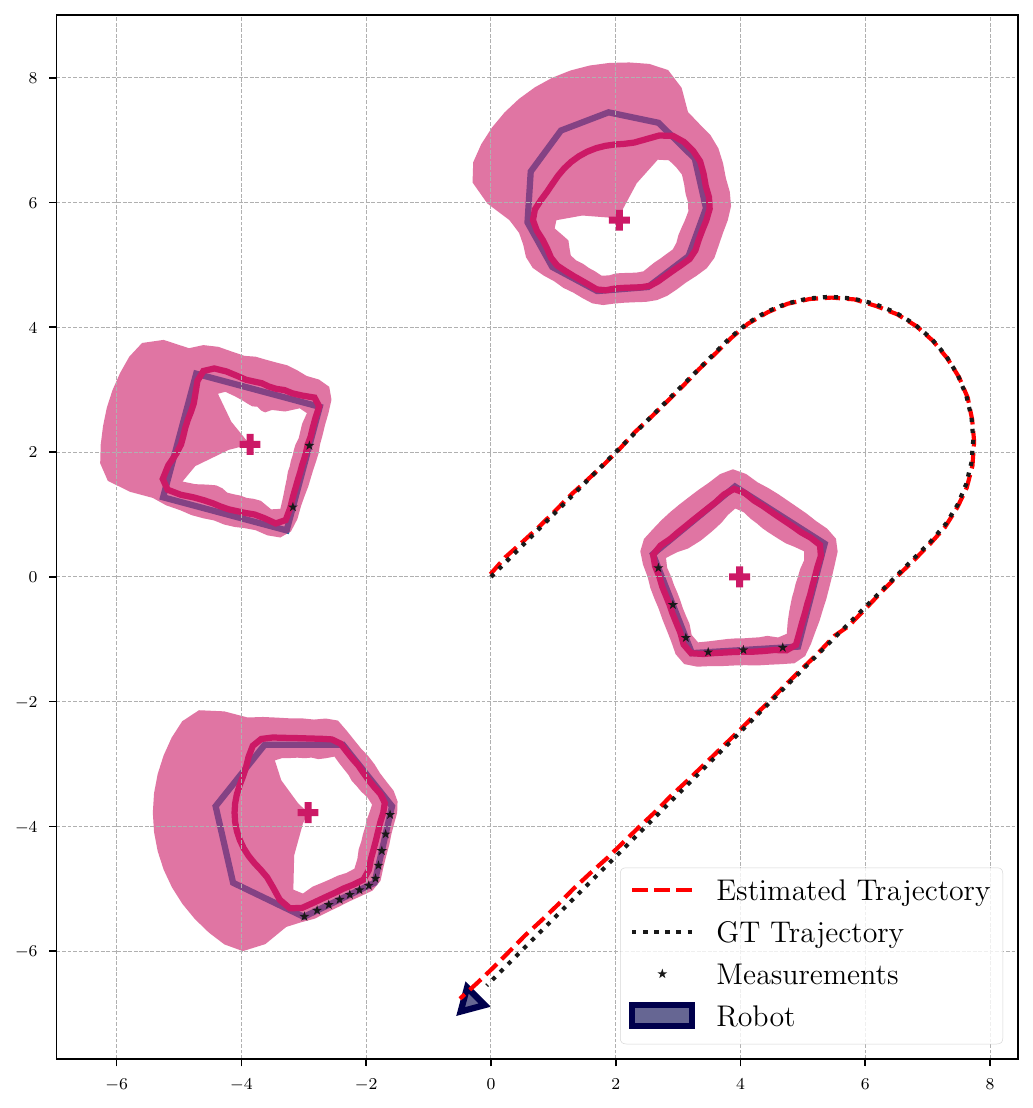}
    \includegraphics[width=.32\textwidth]{figures/sim1/ekf/frame_130.pdf} \\
    \includegraphics[width=.32\textwidth]{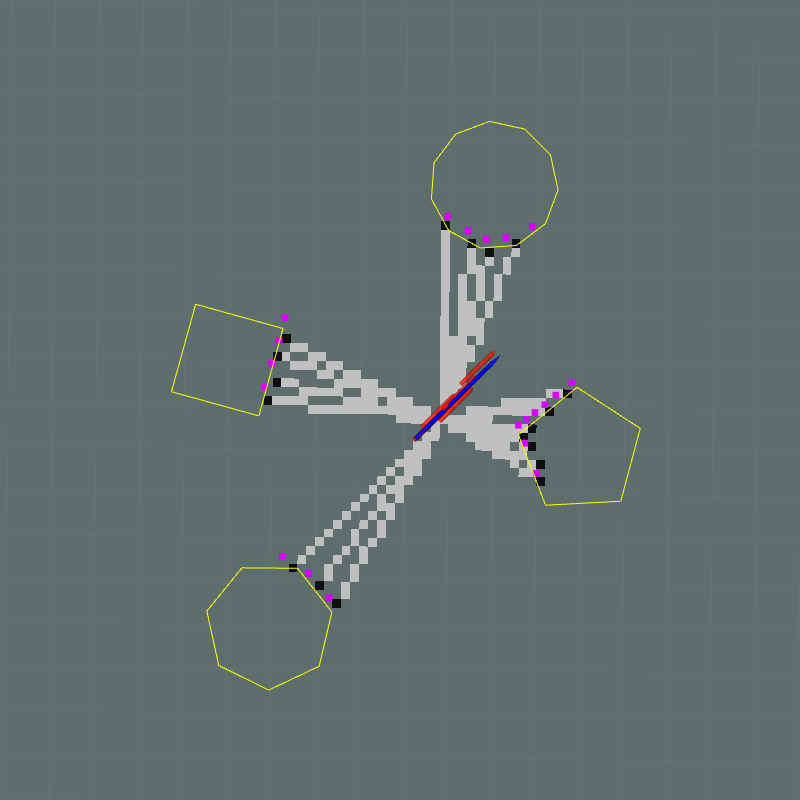}
    \includegraphics[width=.32\textwidth]{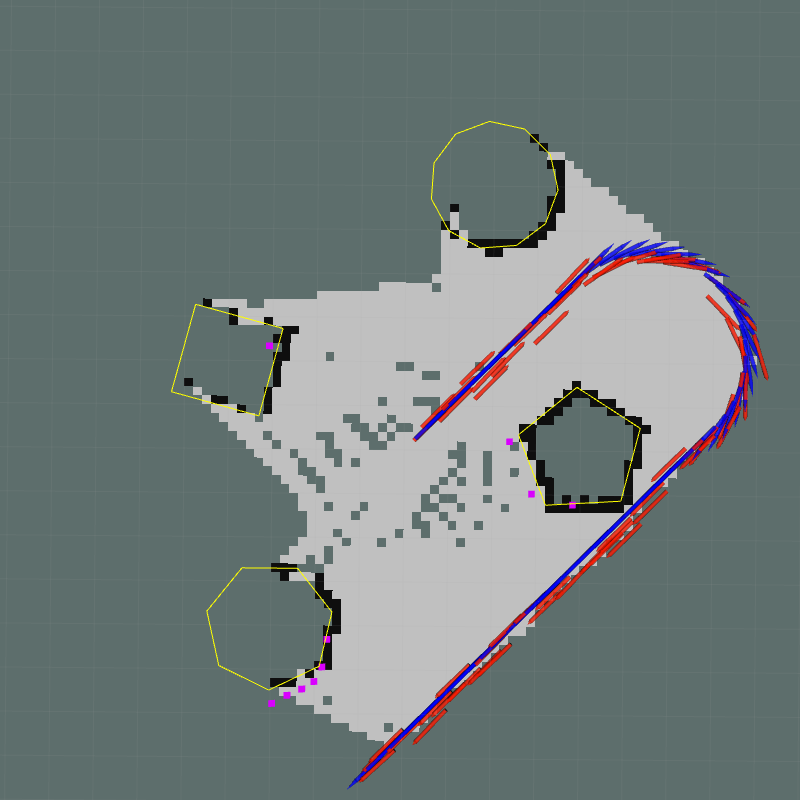}
    \includegraphics[width=.32\textwidth]{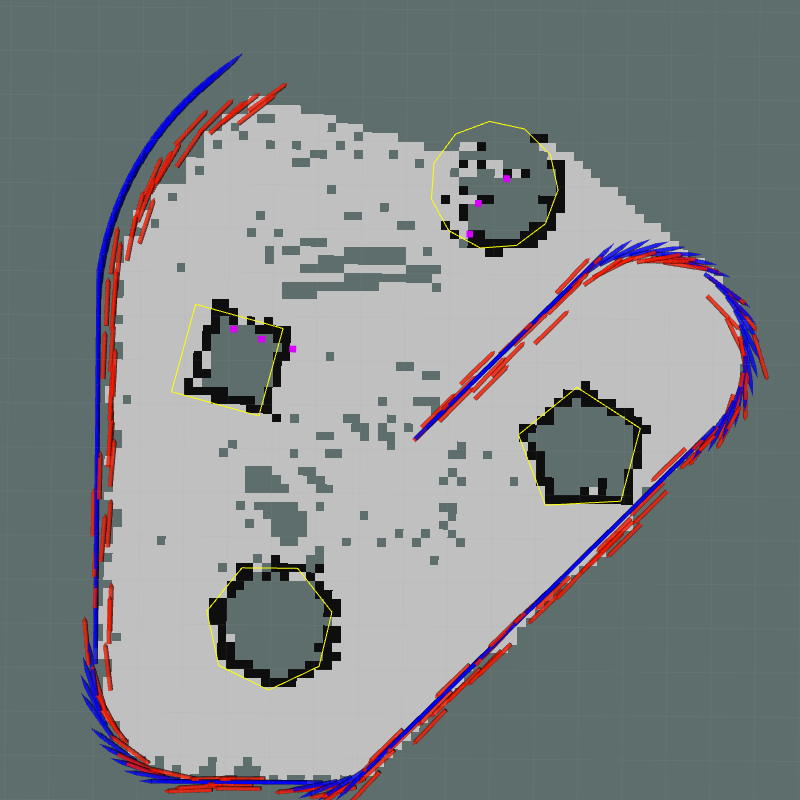}
    \caption{
        Three snapshots of the scenario presented in Section \ref{sec:sim1} where proposed algorithm is presented on the top row and the alternative is presented in the bottom row.
    }
    \label{fig:sim1}
\end{figure*}


\subsection{Simulation Studies} \label{sec:sims}
In this section, we evaluate and compare the proposed approach with the grid based SLAM method introduced in \cite{gmapping}.
The analysis begins with a scenario involving objects with simple geometric shapes.
Then we extend our focus to more complex irregularly shaped objects to assess the performance of both the proposed and the alternative SLAM algorithms in these contexts.
As performance metrics, we use the root mean squared error (RMSE) and the Intersection over Union (IoU) metrics \cite{levandowsky1971distance,padilla2020survey}.

\begin{figure*}[tbp]
    \centering
    \includegraphics[width=.315\textwidth]{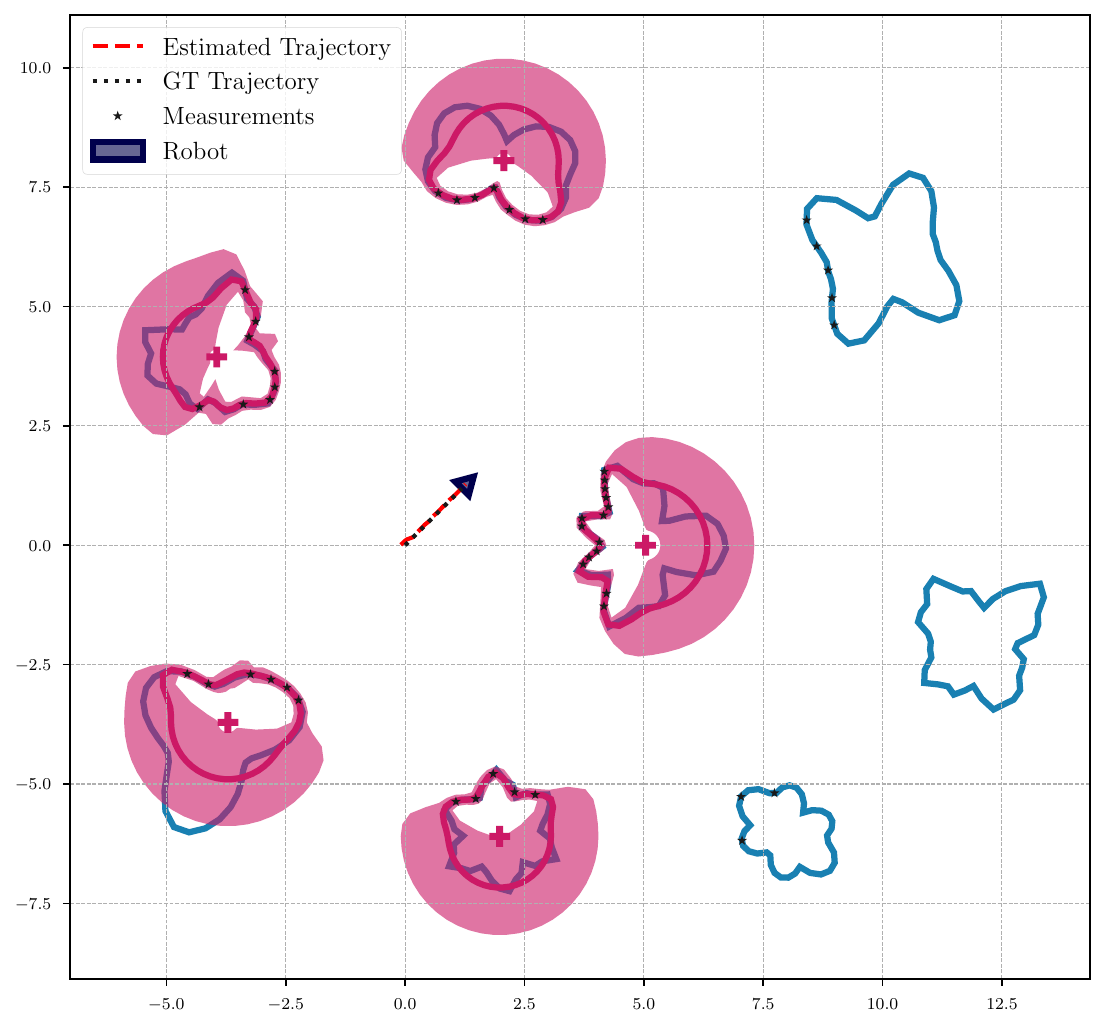}
    \includegraphics[width=.33\textwidth]{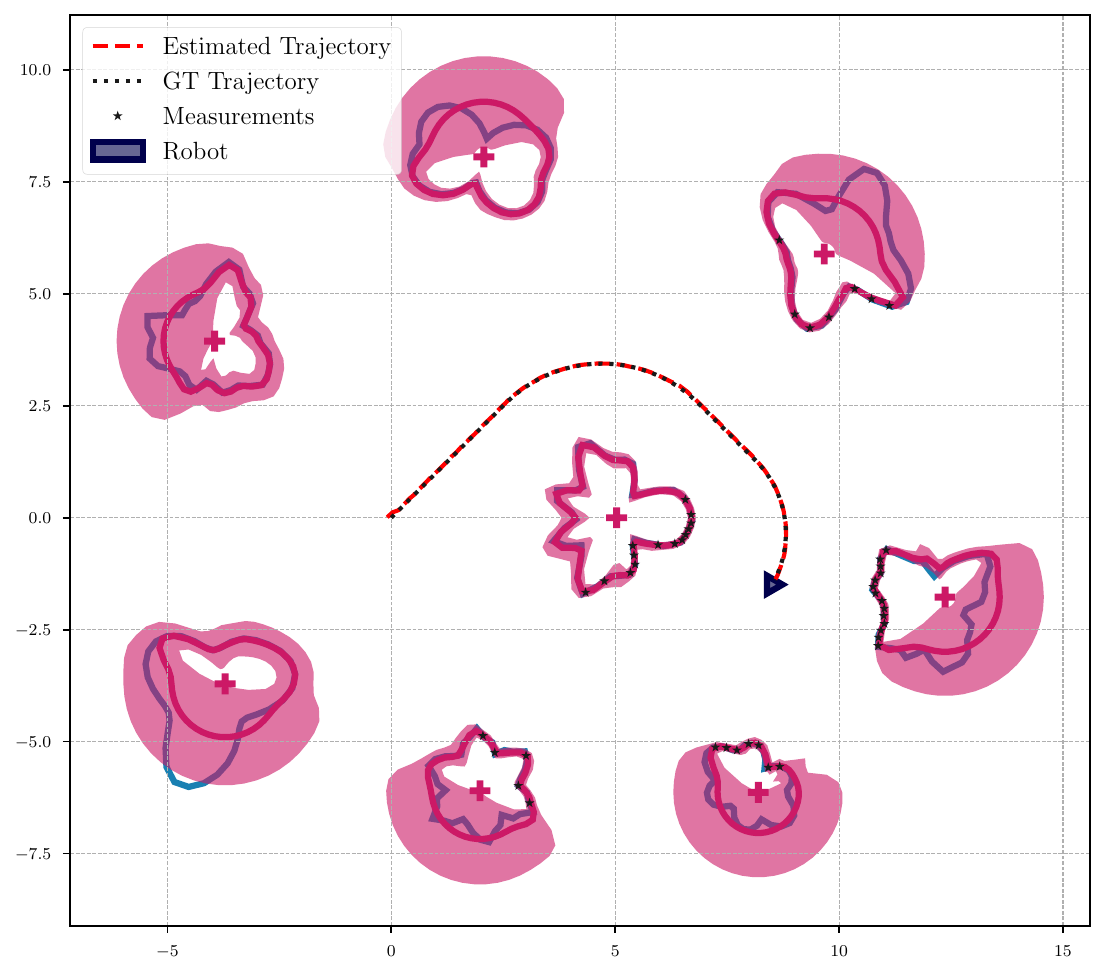}
    \includegraphics[width=.33\textwidth]{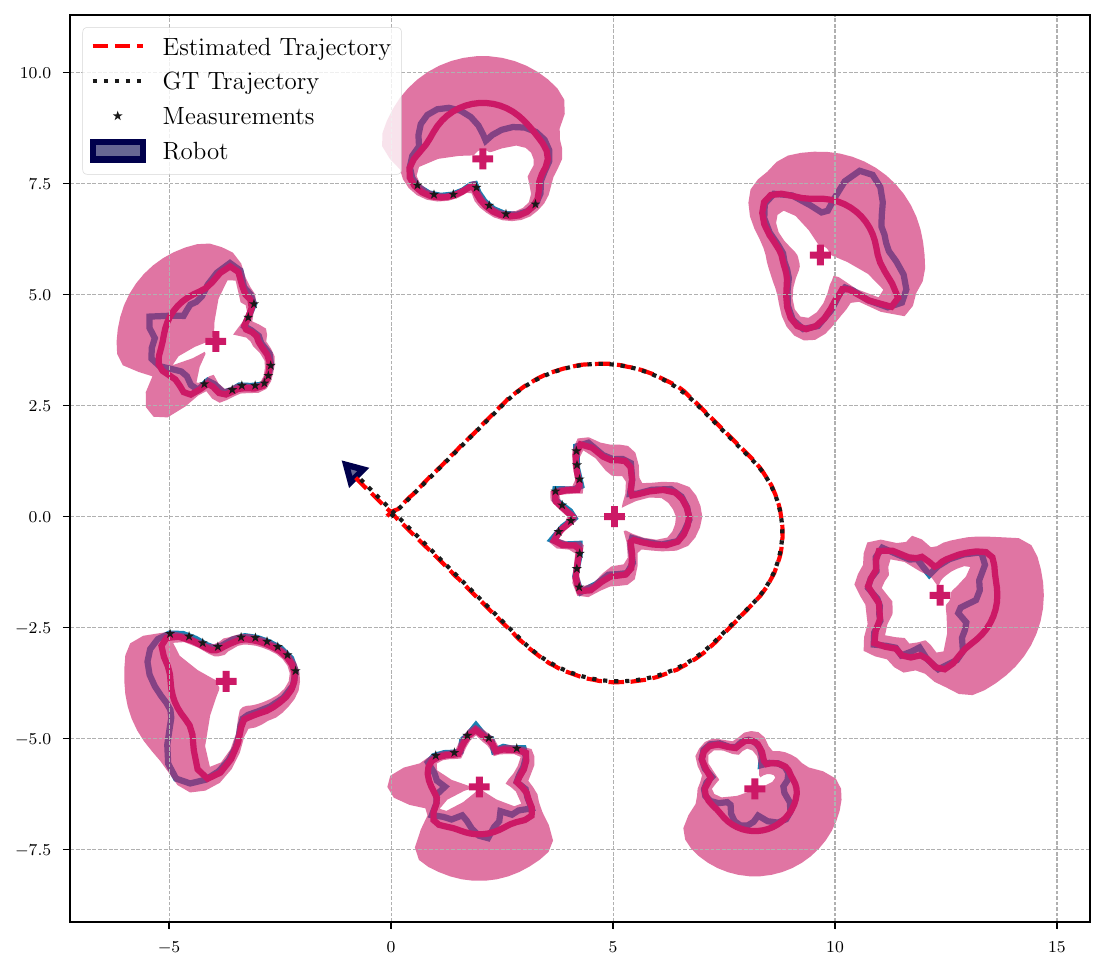} \\
    \includegraphics[width=.32\textwidth]{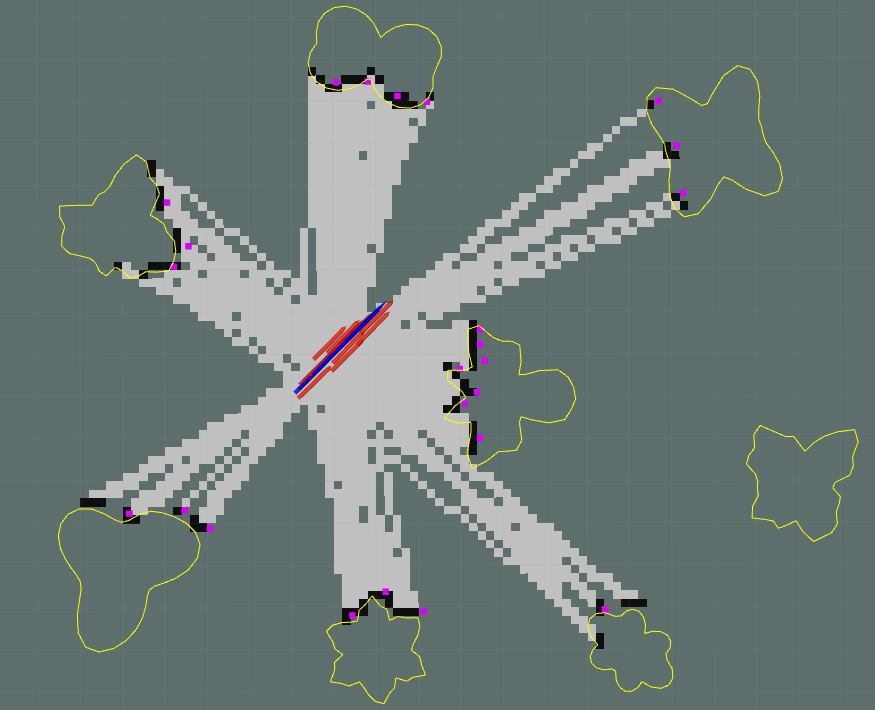}
    \includegraphics[width=.32\textwidth]{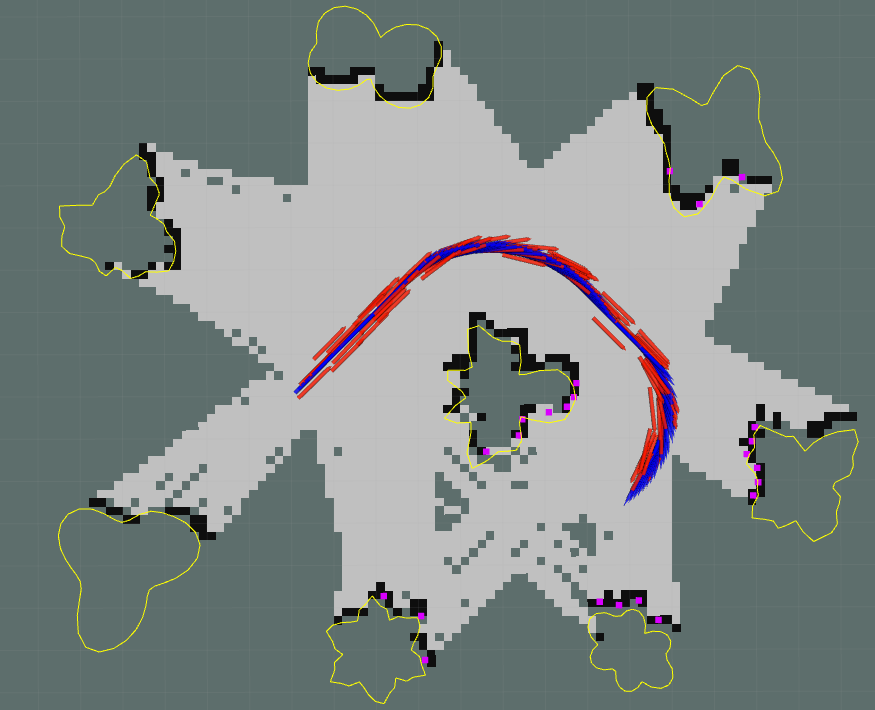}
    \includegraphics[width=.32\textwidth]{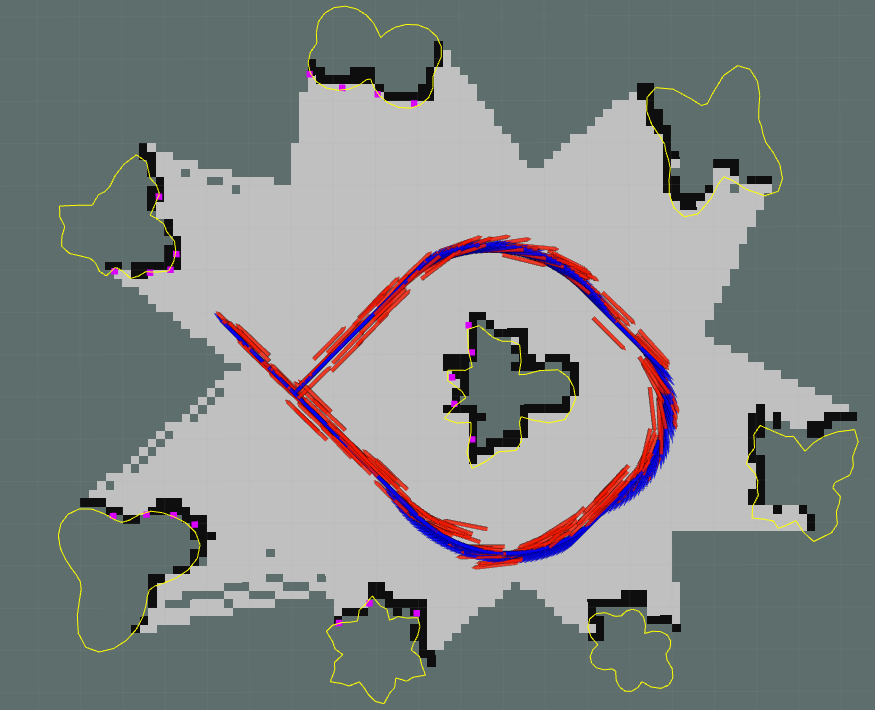}
    \caption{
        Three snapshots of the scenario presented in Section \ref{sec:sim2} where proposed algorithm is presented on the top row and the alternative is presented in the bottom row.
    }
    \label{fig:sim2}
\end{figure*}

\begin{table}[tbp]
    \caption{Time Average RMSE Results for Section \ref{sec:sim1}}
    \centering
    \resizebox{\textwidth}{!}{%
        \begin{tabular}{@{}c|ccc@{}} \label{tab:sim1}
                                                                                            & \multicolumn{1}{c|}{\begin{tabular}[c]{@{}c@{}}\(x\)-coordinate \\ (meters) \end{tabular}} & \multicolumn{1}{c|}{\begin{tabular}[c]{@{}c@{}} \(y\)-coordinate \\ (meters) \end{tabular}} & \multicolumn{1}{c|}{\begin{tabular}[c]{@{}c@{}} Orientation \\ (degrees) \end{tabular}} \\ \midrule
            \begin{tabular}[c]{@{}c@{}} Proposed \\ Method \end{tabular}                    & \( \bm{0.044} \)                                                                           & \( \bm{0.038} \)                                                                            & \( \bm{0.22} \)                                                                         \\ \midrule
            \begin{tabular}[c]{@{}c@{}} Alternative \\ Method \cite{gmapping} \end{tabular} & \( 0.24 \)                                                                                 & \( 0.28 \)                                                                                  & \( 2.69 \)                                                                              \\ \bottomrule
        \end{tabular}%
    }
\end{table}

The simulation studies are executed within a custom ROS package using Python.
For both scenarios, we systematically compare the mapping accuracy and localization errors of the two algorithms.
Throughout all simulations, a robot equipped with a LiDAR and an onboard odometry sensor is simulated to perform SLAM within this environment. The simulated LiDAR sensor has an angular resolution of $3.6^\circ$.

For the benchmark SLAM algorithm, we consider the GMapping algorithm \cite{gmapping, Quigley2009ROSAO}.
We use a grid resolution of \(25 \times 25\) cm for this algorithm, which is a typical choice for environments of this scale, while the number of particles for the benchmark algorithm is set to \(100\), which balances computational efficiency and accuracy.

For object initiation, we use the OPTICS algorithm \cite{ankerst1999optics}, which has proven effective in identifying clusters within datasets characterized by varying densities and selected the number of basis points in all experiments to 50.


\subsubsection{Simple Geometric Shapes} \label{sec:sim1}

\begin{figure*}[tbp]
    \centering
    \includegraphics[width=.49\textwidth]{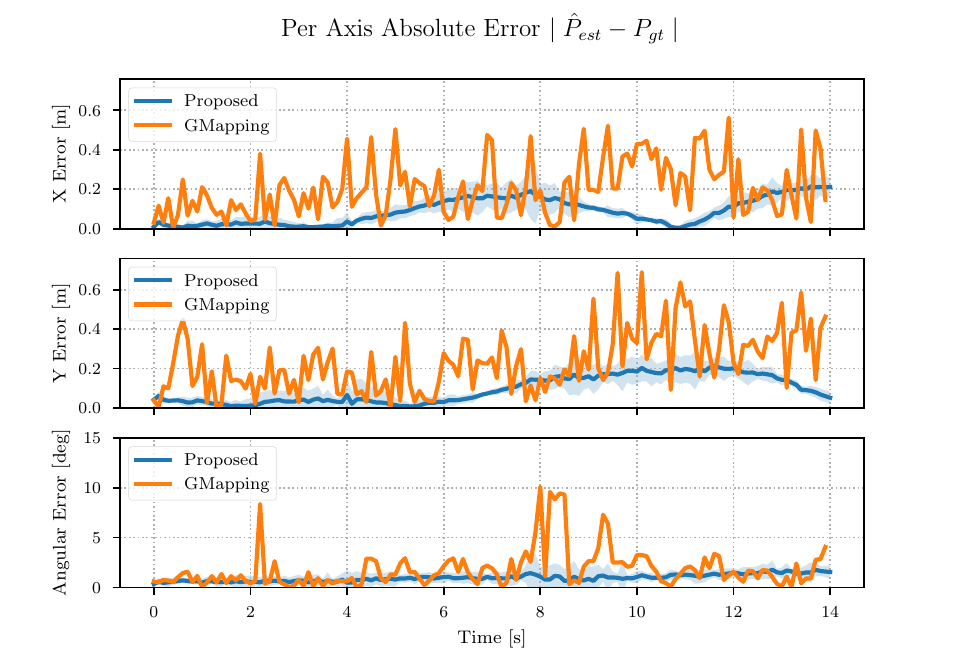}
    \includegraphics[width=.49\textwidth]{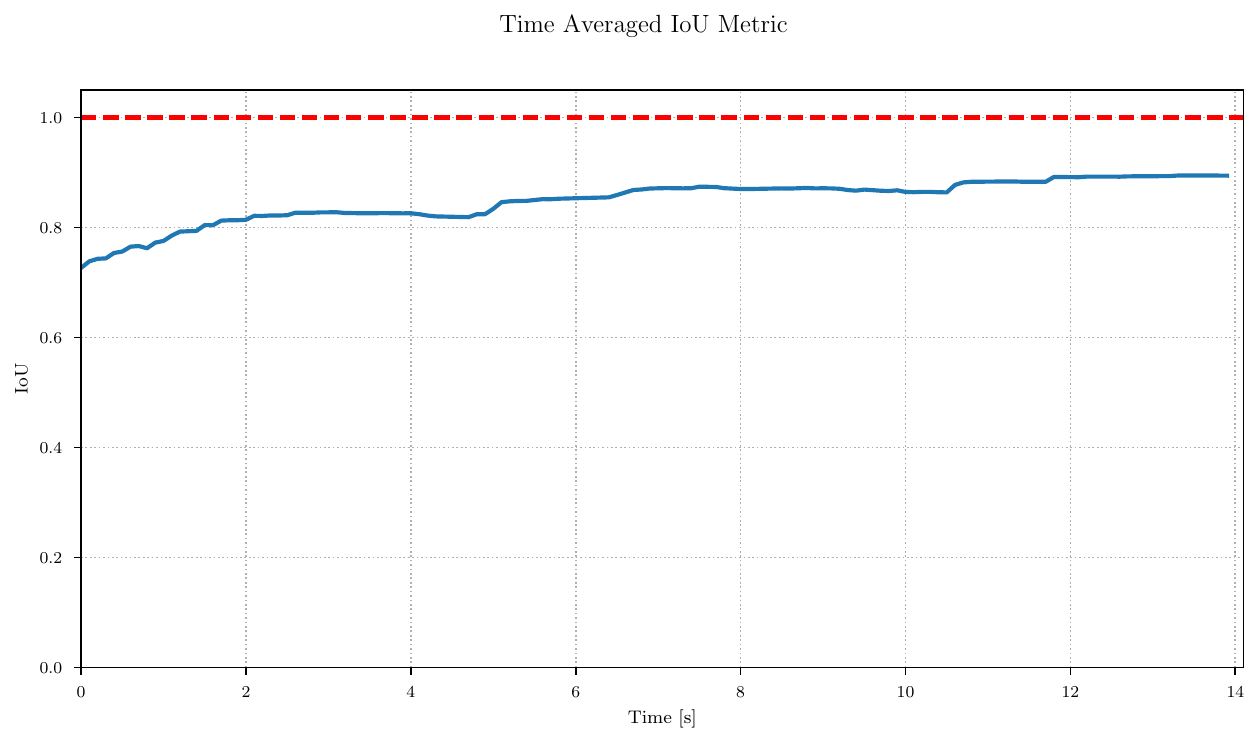}
    \caption{The absolute error plots of the proposed and the alternative algorithm (left) and the IoU plots of the estimates of the proposed algorithm (right) for the scenario in Section \ref{sec:sim1}.
        The shaded areas indicate the experimental \(\% 99\) confidence intervals after \(100\) MC runs.}
    \label{fig:rmse1}
\end{figure*}

In this scenario, the environment contains four map objects, each represented by a regular polygon. The robot is initially positioned at the center of the environment and follows a path consisting of alternating straight lines and circular arcs.
Figure \ref{fig:sim1} depicts snapshots from various stages of the simulation.
The performances of the proposed approach and the alternative approach are presented in Figure \ref{fig:rmse1} and the RMSE values are presented in Table \ref{tab:sim1}.

For the proposed method, the position and orientation RMSE are below \(5\) cm per axis and \(0.22\) degrees in orientation, which are significantly better than the alternative method. The localization performance of the alternative method suffered from the sparsity of the sensor and the distance to the objects, leading to accumulated errors that disturbed the mapping process.
Also, we can see that the IoU metric of the proposed method approaches to the ideal value of \(1\) in a consistent manner, while the localization problems in the grid based approach resulted in an inaccurate map representation.


\subsubsection{Irregular Shapes} \label{sec:sim2}

\begin{figure*}[tbp]
    \centering
    \includegraphics[width=.49\textwidth]{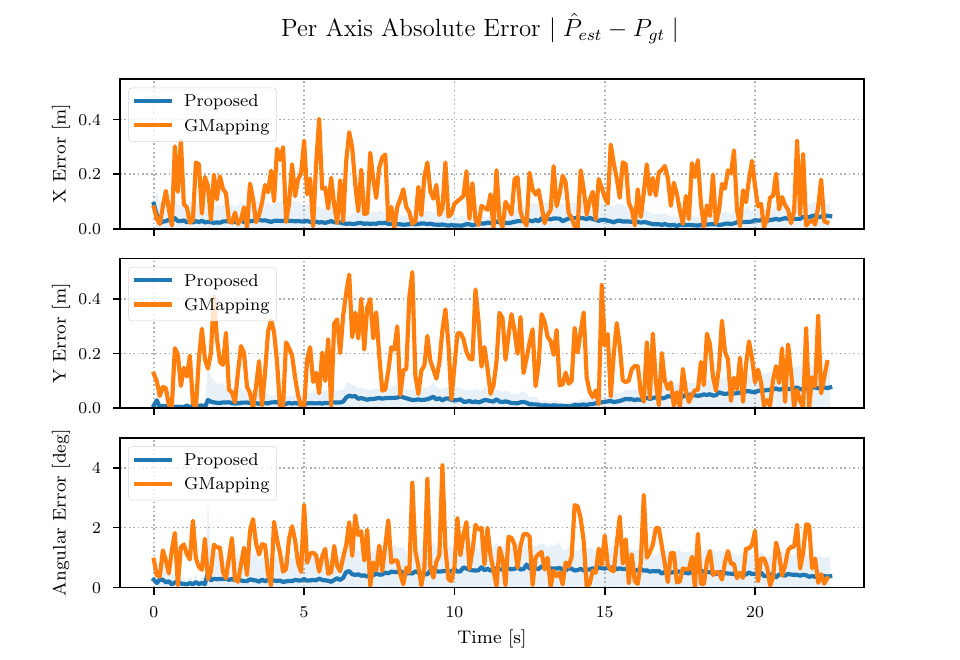}
    \includegraphics[width=.49\textwidth]{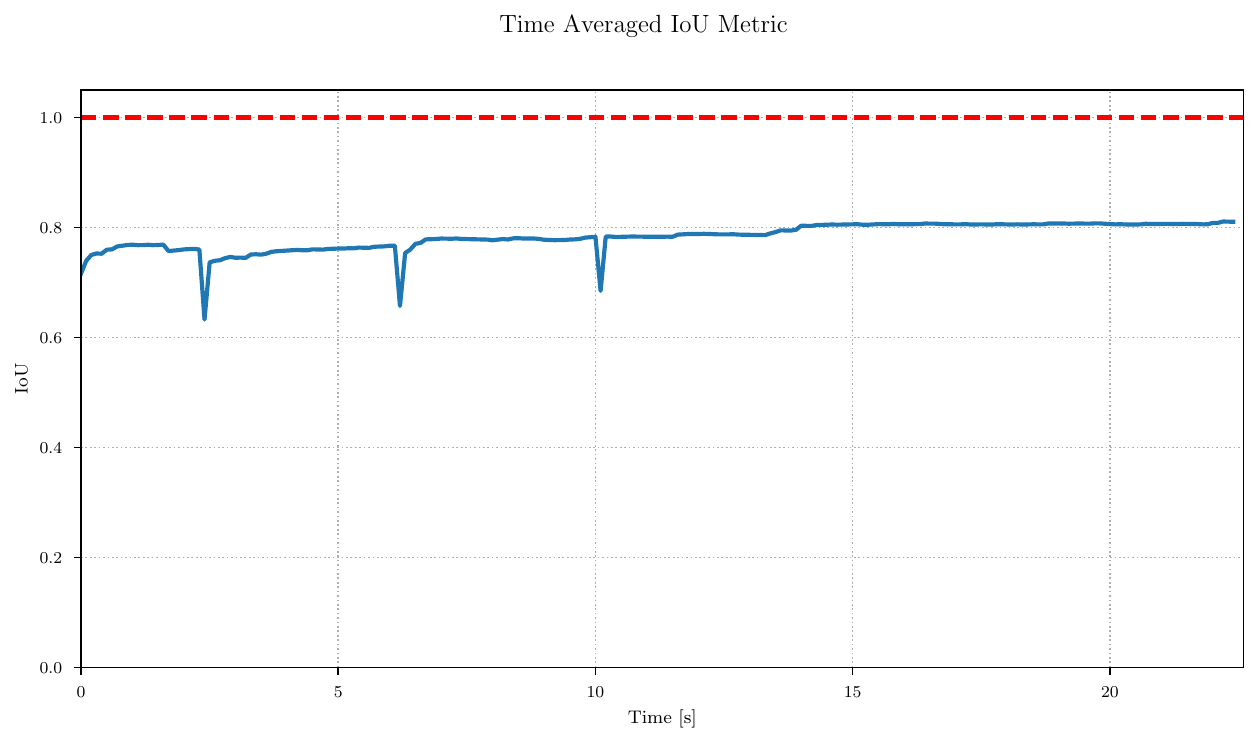}
    \caption{The absolute error plots of the proposed and the alternative algorithm (left) and the IoU plots of the estimates of the proposed algorithm (right) for the scenario in Section \ref{sec:sim2}.
        The shaded areas indicate the \(\% 99\) confidence intervals after \(100\) MC runs.}
    \label{fig:rmse2}
\end{figure*}

This scenario involves eight distinct objects with irregular shapes with smooth contours. The robot again follows a path consisting of alternating straight lines and circular arcs and crosses the starting point again later in the experiment at \(t = 20\) seconds. Initially, not all objects are visible; however, over time they become at least partially visible.
The RMSE values are given in Table \ref{tab:sim2} and the visualizations are given in Figure \ref{fig:sim2} with the error plots per axis in Figure \ref{fig:rmse2}.

With the irregular shapes, the grid based approach performed better compared to the previous scenario, however the proposed method still outperforms the alternative method both in terms of the visual map quality and the localization RMSEs.
Table \ref{tab:sim2} shows that the position RMSE of the proposed algorithm do not exceed \(10\) centimeters in any direction, with orientation errors remaining below \(1.48\) degrees. Also, an important aspect of the proposed algorithm lies in its ability to compansate for the drifts without explicit loop closure detection and processing, as at \(t = 20\) seconds, the proposed method still has a low drift in pose estimation visible in Figure \ref{fig:rmse2}.
The proposed method experiences occasional IoU drops due to the initialization of the new objects, but this drop is quickly compensated when the new object is estimated more accurately over time and the IoU metric remains above \(0.60\) throughout the experiment.

\begin{table}[tbp]
    \caption{Time Average RMSE Results for Section \ref{sec:sim2}}
    \centering
    \resizebox{\textwidth}{!}{%
        \begin{tabular}{@{}c|ccc@{}} \label{tab:sim2}
                                                                                            & \multicolumn{1}{c|}{\begin{tabular}[c]{@{}c@{}}\(x\)-coordinate \\ (meters) \end{tabular}} & \multicolumn{1}{c|}{\begin{tabular}[c]{@{}c@{}} \(y\)-coordinate \\ (meters) \end{tabular}} & \multicolumn{1}{c|}{\begin{tabular}[c]{@{}c@{}} Orientation \\ (degrees) \end{tabular}} \\ \midrule
            \begin{tabular}[c]{@{}c@{}} Proposed \\ Method \end{tabular}                    & \( \bm{0.057} \)                                                                           & \( \bm{0.093} \)                                                                            & \( 1.48 \)                                                                              \\ \midrule
            \begin{tabular}[c]{@{}c@{}} Alternative \\ Method  \cite{gmapping}\end{tabular} & \( 0.131 \)                                                                                & \( 0.198 \)                                                                                 & \( \bm{1.26} \)                                                                         \\ \bottomrule
        \end{tabular}%
    }
\end{table}

\subsection{Real World Scenario}
\begin{figure}[tbp]
    \centering
    \includegraphics[width=.9\textwidth]{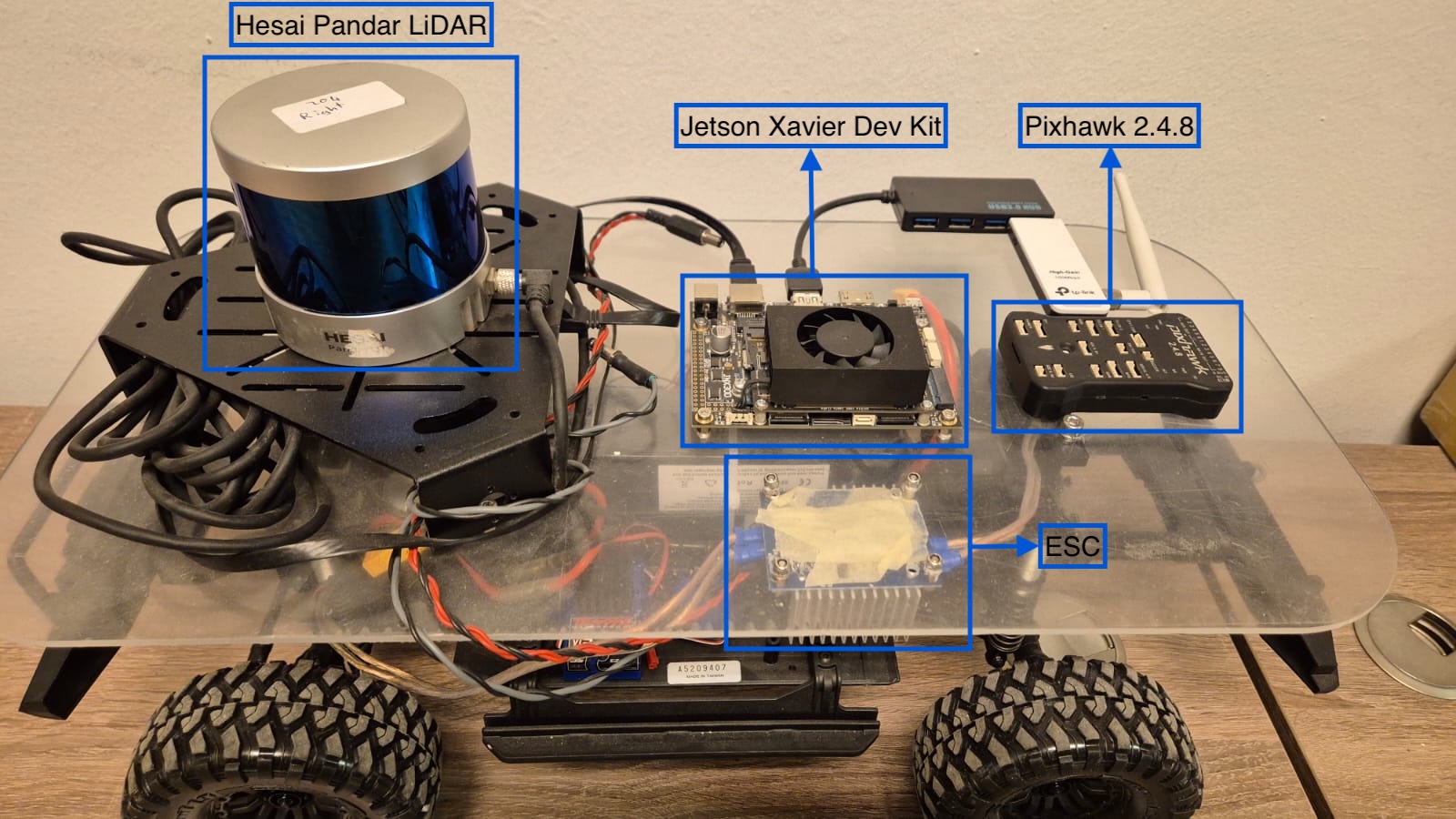}
    \caption{Custom mobile platform setup for the real world scenario}
    \label{fig:exp_setup}
\end{figure}

In this section, we assess the performance of our proposed algorithm in a real world environment.
The experiment was conducted in a parking lot, where we aimed to estimate the shapes and positions of the three cars present.
Data was collected using a custom mobile platform equipped with a Hesai PandarQT LiDAR \cite{hesai_pandarQT} sensor mounted on top where a single LiDAR channel is used to generate a 2D measurements, along with a Pixhawk 2.4.8 \cite{pixhawk248} module providing IMU data, see Figure \ref{fig:exp_setup}.

\begin{figure*}[tbp]
    \centering
    \begin{subfigure}[b]{0.32\textwidth}
        \centering
        \includegraphics[width=\textwidth]{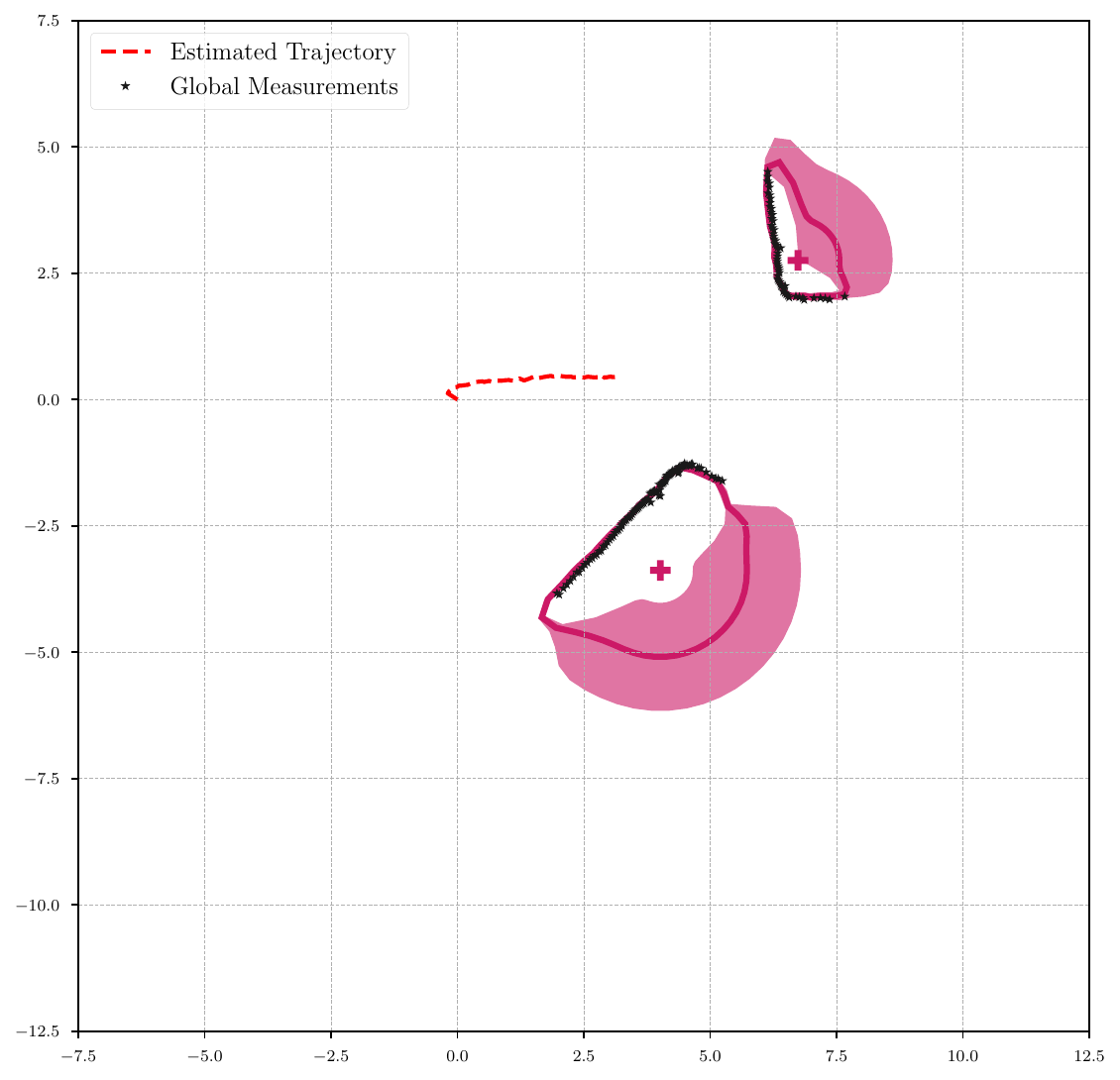}
    \end{subfigure}%
    \hfill
    \begin{subfigure}[b]{0.32\textwidth}
        \centering
        \includegraphics[width=\textwidth]{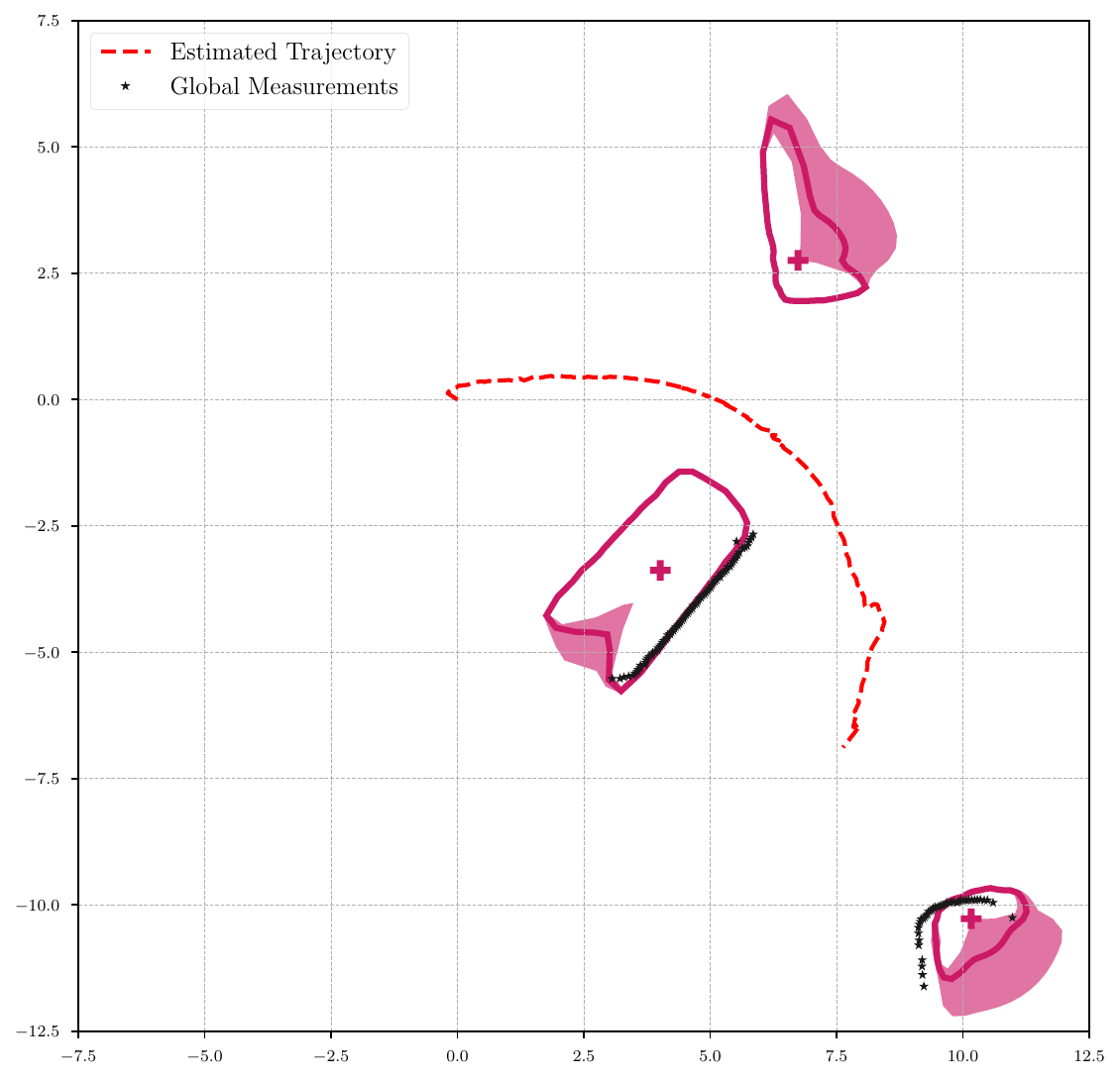}
    \end{subfigure}%
    \hfill
    \begin{subfigure}[b]{0.32\textwidth}
        \centering
        \includegraphics[width=\textwidth]{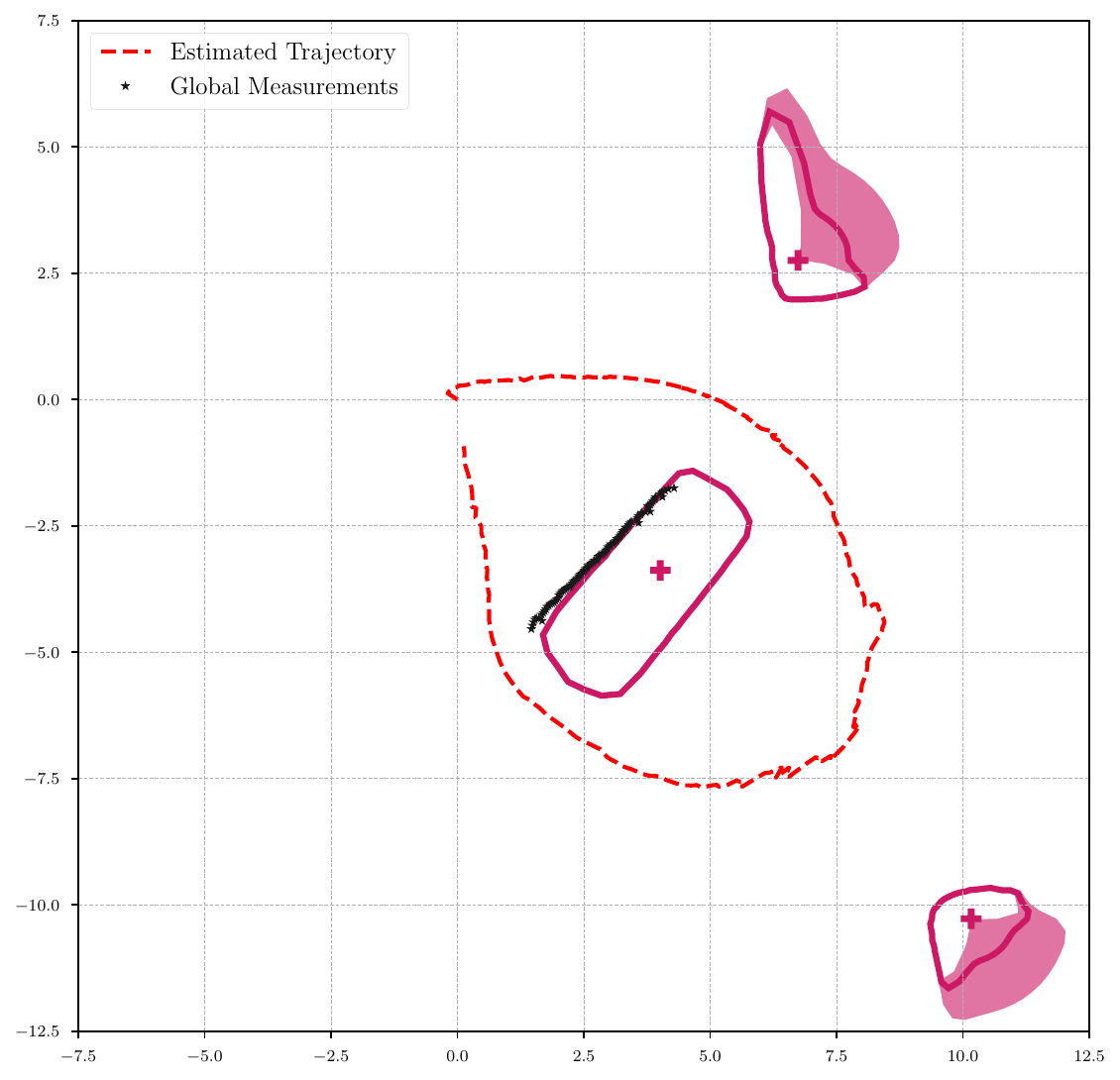}
    \end{subfigure}

    \vspace{0.5em}

    \begin{subfigure}[b]{0.33\textwidth}
        \centering
        \includegraphics[width=\textwidth]{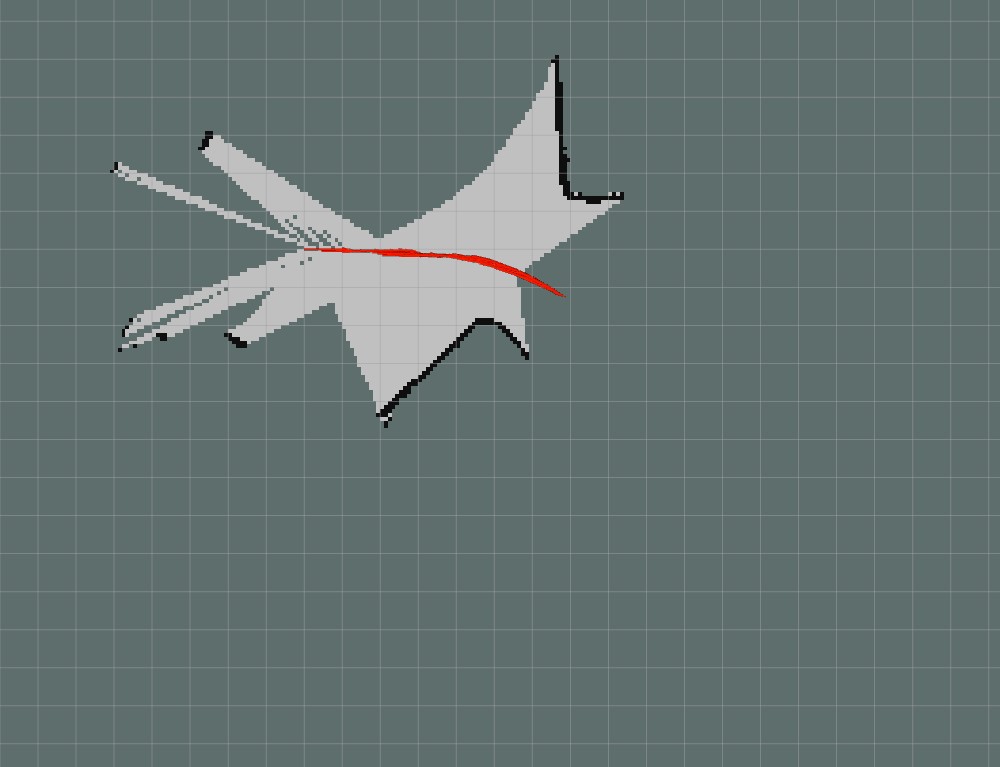}
    \end{subfigure}%
    \hfill
    \begin{subfigure}[b]{0.31\textwidth}
        \centering
        \includegraphics[width=\textwidth]{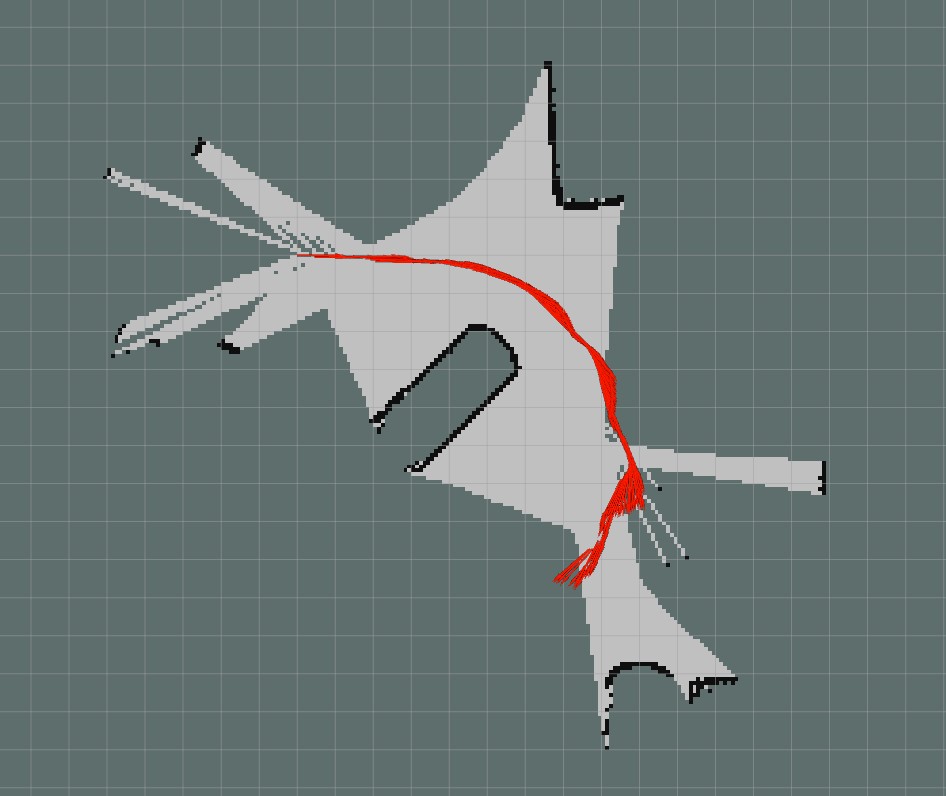}
    \end{subfigure}%
    \hfill
    \begin{subfigure}[b]{0.3\textwidth}
        \centering
        \includegraphics[width=\textwidth]{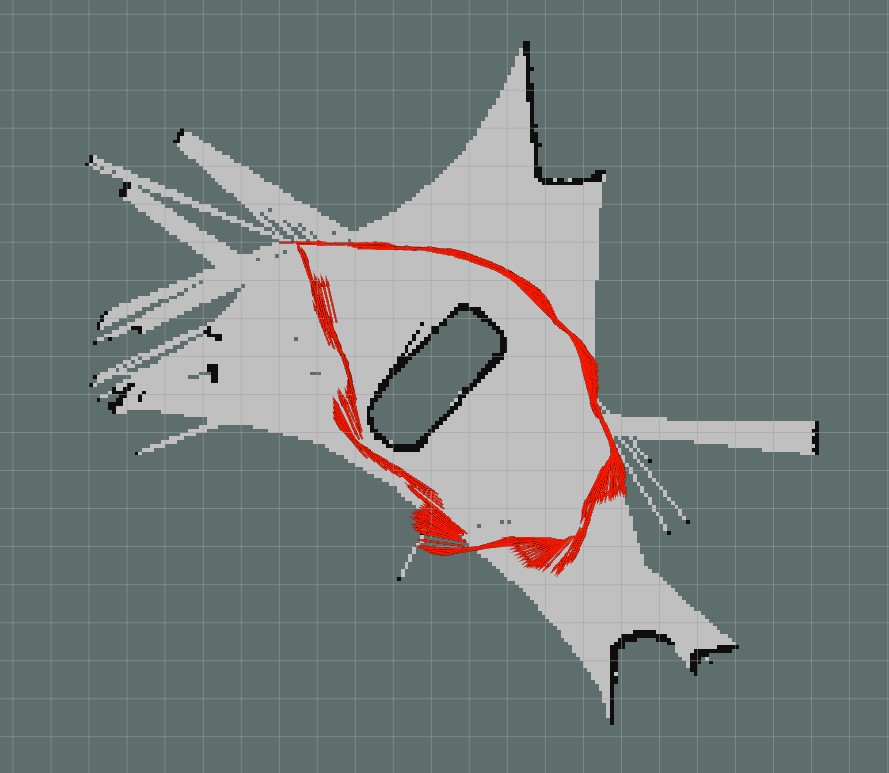}
    \end{subfigure}

    \caption{
        Three snapshots of the real world scenario where proposed algorithm is presented on the top row and the alternative is presented in the bottom row.
    }
    \label{fig:real_data}
\end{figure*}

Three snapshots from various stages of the experiment is presented on Figure \ref{fig:real_data}.
Our approach was able to correctly estimate the shapes and positions of the three cars, especially the car in the middle which was explored in its entirety.
Visual comparisons of the maps suggest that the likelihood based association approach correctly rejected the outlier measurements whereas the grid based approach was affected negatively by the outliers, which undermine the map and localization accuracy.

%% file: sections/discussion.tex
\section{Discussion}\label{sec:discussion}

Our approach offers shape estimates with a quantifiable uncertainty bound, while the alternative method \cite{gmapping} provides grid occupancy without uncertainty quantification.
This distinction is critical for applications requiring probabilistic guarantees in map representation.
The performance of our representation can be seen in Figure \ref{fig:sim1}, where the alternative approach fails to correctly localize in the final frame.
Since lidar data is sparse in both experiments, outliers in the laser scan data can easily interfere with the scan matching process, leading to distortions in the grid map.
Our method, however, implements a per object maximum likelihood based approach for data association, which facilitates the rejection of outlier measurements.
This robustness is particularly valuable in environments where sensor noise and data anomalies are common.

Considering the map occupancy representations, unoccupied cells tend to remain undetected by the alternative approach for extended periods, as illustrated in Figures \ref{fig:sim1} and \ref{fig:sim2}.
By contrast, our approach can provide occupancy representation probabilistically for any point on the map, thanks to the uncertainty bounds associated with map objects.
This capability clearly shows the regions where the robot has not yet explored in the environment, which could be useful for high level tasks such as path planning, decision making and exploration.


In the real data scenario, the proposed approach successfully reconstructed the visible parts of the objects.
The uncertainty bounds clearly indicate the regions where the robot has not yet explored in the environment as seen in Figure \ref{fig:real_data}, while the alternative method fails to detect outlier measurements and produce significantly noisier spatial representations.
This disparity stems directly from our object centric data association framework, which enables systematic outlier rejection through maximum likelihood validation, a capability absent in grid based approaches.
This combination of precision and practicality addresses a critical gap in mobile robotics: the need for statistically rigorous maps that balance detail with computational constraints.
While our experiments focused on structured environments, the proposed uncertainty quantification framework suggests broader applicability in cluttered outdoor settings where transient occlusions are common.
Additionally, the robustness of our method to sensor noise and data anomalies is particularly valuable in environments where such disturbances are common, which is a sign of reliable performance across diverse scenarios.

An important note is about loop closure capabilities of our algorithm.
Without explicitly calculating whether the robot revisits the same position, we achieve low drift localization thanks to per object measurement association.
The sufficient statistics embedded in each object's representation contain all necessary information from past updates, which enables the robot to maintain accurate localization.

%% file: sections/conclusion.tex
\section{Conclusion}
In this paper, we introduce a novel online SLAM framework that leverages GP based object representations to improve mapping and localization accuracy.
Our approach models map objects as static 2D volumetric shapes, accommodating a wide range of geometric configurations to accurately reflect the diversity of real world environments.
We parameterize the contours of map objects as radial functions, employing star-convex sets as the underlying mathematical representation.
By utilizing a recursive GP model, we are able to estimate these radial functions in real time within a recursive Bayesian estimation framework.
The GP based contour model is particularly advantageous as it exploits the inherent spatial correlations between pairs of points on the contour, enabling the system to infer the statistical properties of unobserved segments of object contours.
This capability is of paramount importance for tasks such as predicting object shapes with quantified uncertainties, which is crucial for applications in navigation, data association, and exploration.
The GP model's ability to provide confidence bounds for object contours represents a significant advancement in the accuracy and reliability of SLAM systems, offering new opportunities for improving autonomous systems' interaction with their environments.

Our approach represents a significant extension to typical LiDAR SLAM methods which often rely on grid mapping and point cloud registration techniques to represent objects, our method provides a more compact and efficient representation by modeling objects on a per object basis.
This results in a memory efficient representation, as each object is stored as a few statistically related variables rather than an extensive collection of spatially independent variables.
By reducing the memory requirements associated with SLAM, our approach not only enhances the scalability of SLAM systems but also improves their computational efficiency, making it feasible to deploy SLAM with limited sensors or on platforms with limited computational capabilities. Also, this extension of the typical SLAM framework to object centric mapping does not conflict with existing methods since we use a Bayesian framework for our system. With this framework, it is possible to use a different SLAM pipeline with different characteristics, for example a grid based method for the non star-convex parts of the environment and our method for the star-convex parts. As an alternative, a detector can be used to detect some objects of interest and our method can be used for the detected parts of the environment. Within this hybrid framework, our system can provide Bayesian estimates for the map and the pose of the robot to be fused with the other SLAM systems.

Looking forward, we will focus on extending our approach to accommodate non-star-convex shapes, thereby further enhancing the flexibility and accuracy of map representations.
This extension will be particularly important for capturing more complex object geometries that are frequently encountered in real world environments.
Additionally, from an application perspective, we plan to enhance the computational efficiency of our method by integrating advanced EKF models that have been previously explored in the literature \cite{julier2007using, paz2008divide}.
These models hold the potential to further optimize the performance of our SLAM framework, making it more suitable for real time applications.
Moreover, we aim to adapt our approach for use in 3D SLAM, which will significantly broaden its applicability to a wider range of autonomous systems and environments.
We anticipate that our framework will contribute to the ongoing evolution of SLAM research ultimately leading to the development of more sophisticated and capable autonomous systems in the future.


%



%% file: sections/appendix.tex
\appendices

\section{IEKF Update Equations} \label{ap:iekf}
Consider a state space model similar to \eqref{eq:augmented_full_state_space} as
\begin{subequations} \label{eq:augmented_full_state_space_linear}
    \begin{align}
        \bx_{k+1} & = g_k(\bx_k) + \bu_k + \bw_k, \quad & \bw_k \sim \N(\bm{0}, \bQ_k), \\
        \bz_k     & = h_k(\bx_k) + \bv_k, \quad         & \bv_k \sim \N(\bm{0}, \bR_k), \\
        \bx_0     & \sim \N(\hat{\bx}_0, \bSig_0).
    \end{align}
\end{subequations}
For this state space model, an iterated Kalman filter (IEKF) can be used to obtain solutions for the state estimation problem \cite{havlik2015performance}.
The time update for the IEKF formulation is the same as the standard EKF formulation, however, the measurement update is modified slightly.
At each iteration the linearization point, for the iteration number \(\ell\), is modified as
\begin{subequations}
    \begin{align}
        \bH_k^\ell           & = \frac{d h_k(\bx)}{d \bx} \bigg \rvert_{\bx = \hat{\bx}_k^\ell},                                                                       \\
        \bK_k^\ell           & =  \bSig_{k \mid k-1} (\bH_k^\ell)^\tr (\bH_k^\ell \bSig_{k \mid k-1} (\bH_k^\ell)^\tr +\bR_k)^{-1},                                    \\
        \hat{\bx}_k^{\ell+1} & = \hat{\bx}_{k\mid k-1} + \bK_k^\ell \bigl(\bz_k - h_k(\hat{\bx}_k^\ell) - \bH_k^\ell(\hat{\bx}_{k \mid k-1} - \hat{\bx}_k^\ell)\bigr).
    \end{align}
\end{subequations}
The estimated state is updated at the final linearization point as
\begin{subequations}
    \begin{align}
        \hat{\bx}_{k \mid k} & = \hat{\bx}_k^\ell,                                             \\
        \bSig_{k \mid k}     & = \bSig_{k \mid k-1} -  \bK_k^\ell\bH_k^\ell\bSig_{k \mid k-1}.
    \end{align}
\end{subequations}

\section{Calculation of Derivatives} \label{ap:jacobians}
For EKF implementation, Jacobians of the state space equations is needed.
Jacobian of the proposed measurement model can be defined as
\scriptsize
\begin{align}
    \bH_k                                    & \triangleq \nabla h_k(\hat{\bx}_{k \mid k-1}) \nonumber                                                                                                                                                                                                                                                           \\
                                             & =  \mtx{\nabla h_k^1(\bx)^\tr                                                                                                                              & \nabla h_k^2(\bx)^\tr             & \dots                             & \nabla h_k^N(\bx)^\tr}^\tr \bigg \rvert_{\bx = \hat{\bx}_{k \mid k-1}},      \\
    \nabla h^i_k(\hat{\bx}_{k \mid k-1})     & =  \text{blkdiag} \Bigl(\nabla h_k^{i,1}(\bx), \nabla h_k^{i,2}(\bx), \dots, \nabla h_k^{i,m_k^i}(\bx) \Bigr) \bigg \rvert_{\bx = \hat{\bx}_{k \mid k-1}},                                                                                                                                                        \\
    \nabla h_k^{i,j}(\hat{\bx}_{k \mid k-1}) & = \mtx{ \frac{\partial h^{i,j}(\bx)}{\partial \bx^{i,c}}                                                                                                   & \frac{d h^{i,j}(\bx)}{d\bx^{i,f}} & \frac{d h^{i,j}(\bx)}{d\bx^{r,p}} & \frac{d h^{i,j}(\bx)}{d\phi} } \bigg \rvert_{\bx = \hat{\bx}_{k \mid k-1 }}.
\end{align}
\normalsize
There are four components needed to calculate the Jacobian for each object, which can be obtained using the chain rule
\scriptsize
\begin{subequations}
    \begin{align}
        \frac{\partial h^{i,j}(\bx)}{ \partial \bx^{i,c}} & = \bT(\phi)^\tr \biggl( \mathbb{I}_2 + \frac{\partial \bp(\theta^{i,j})}{ \partial \bx^{i,c}} \bH^f(\theta^{i,j}) \nonumber                         \\
                                                          & + \bp(\theta^{i,j}) \frac{\partial \bH^f(\theta^{i,j})}{\partial \theta^{i,j}} \frac{\partial \theta^{i,j}}{\partial \bx^{i,c}}  \bx^{i,f} \biggr), \\
        \frac{\partial h^{i,j}(\bx)}{\partial \bx^{i,f}}  & = \bT(\phi)^\tr \bp(\theta^{i,j}) \bH_k^f(\theta^{i,j}),                                                                                            \\
        \frac{\partial h^{i,j}(\bx)}{\partial \bx^{r,p}}  & = -\frac{\partial h_k^{i,j}(\bx)}{ \partial \bx^{i,c}},                                                                                             \\
        \frac{\partial h^{i,j}(\bx)}{\partial \phi}       & = \frac{\partial \bT(\phi)^\tr}{\partial \phi}
        \biggl(\bx^{i,c} + \bp(\theta^{i,j}) \bH^f(\theta^{i,j})\bx^{i,f} - \bx^{r, p}\biggr) \nonumber                                                                                                         \\
                                                          & -\bT(\phi)^\tr \biggl( \frac{\partial \bp(\theta^{i,j})}{\partial \phi}  \bH^f(\theta^{i,j}) \bx^{i,f} \nonumber                                    \\
                                                          & + \bp(\theta^{i,j}) \frac{\partial \bH^f(\theta^{i,j})}{\partial \theta^{i,j}} \frac{\partial \theta^{i,j}}{\partial \phi}  \bx^{i,f} \biggr),
    \end{align}
\end{subequations}
\normalsize
where
\scriptsize
\begin{subequations}
    \begin{align}
        \frac{\partial \bp(\theta^{i,j})}{ \partial \bx^{i,c}}                   & =   \frac{ (\bT(\phi) \bz^{i,j} + \bx^{r,p} - \bx^{i,c})(\bT(\phi) \bz^{i,j} + \bx^{r,p} - \bx^{i,c})^\tr }{|| \bT(\phi) \bz^{i,j} + \bx^{r,p} - \bx^{i,c}||^3}                  \nonumber                                                                                                                      \\
                                                                                 & - \frac{\mathbb{I}_{2}}{|| \bT(\phi) \bz^{i,j} + \bx^{r,p} - \bx^{i,c}||},                                                                                                                                                                                                                                      \\
        \frac{\partial \bp_k^{i,j}}{ \partial \phi}                              & = \frac{\partial \bp(\theta^{i,j})}{\partial \bx^{i,c}} \frac{\partial \bT(\phi)}{\partial \phi} \bz_k^{i,j},                                                                                                                                                                                                   \\
        \frac{\partial \theta^{i,j}}{ \partial \bx^{i,c}}                        & = \frac{ \biggl( \mtx{-1                                                                                                                                                                   & 0                                                                                                                  \\ 0 & 1} (\bT(\phi) \bz^{i,j} + \bx^{r,p}) - \bx^{i,c}\biggr)^\tr }{||\bT(\phi) \bz^{i,j} + \bx^{r,p} - \bx^{i,c}||^2}  \mtx{0                                                                                                                            & 1                                                                                                                  \\ 1 & 0},\\
        \frac{\partial \theta^{i,j}}{ \partial \phi}                             & = \frac{\partial  \theta^{i,j}}{\partial \bx^{i,c}}  \frac{\partial \bT(\phi)}{\partial \phi} \bz^{i,j}   ,                                                                                                                                                                                                     \\
        \frac{\partial \bT(\phi)}{\partial \phi}                                 & = \mtx{-\sin(\phi)                                                                                                                                                                         & -\cos(\phi)                                                                                                        \\ \cos(\phi) & -\sin(\phi)}, \\
        \frac{\partial \bH^f(\theta^{i,j})}{ \partial \theta^{i,j}}              & = \frac{\partial \bK(\theta^{i,j}, \bar\btheta)}{\partial \theta^{i,j}} \bK(\bar{\btheta}, \bar{\btheta})^{-1},                                                                                                                                                                                                 \\
        \frac{\partial \bK(\theta^{i,j}, \bar{\btheta})}{\partial \theta^{i,j} } & = \frac{\partial \mtx{  \bar{k}(\theta^{i,j}, \bar{\theta}_1)                                                                                                                              & \bar{k}(\theta^{i,j}, \bar{\theta}_2) & \dots & \ \bar{k}(\theta^{i,j}, \bar{\theta}_M)} }{\partial \theta^{i,j}}, \\
        \frac{\partial \bar{k}(\theta, \theta')}{\partial \theta}                & = - \frac{\sigma_f^2\sin(\theta - \theta')}{l^2}  \exp \biggl({-\frac{2\sin(|\theta - \theta'|)^2}{l^2}} \biggr).
    \end{align}
\end{subequations}
\normalsize

%% file: sections/references.tex

\bibliographystyle{IEEEtran}
\bibliography{references}
%

